\documentclass[runningheads]{llncs}

\usepackage{pifont}
\usepackage{enumitem}
\usepackage{multirow}
\usepackage{microtype}
\usepackage{rotating}
\usepackage[hyphens]{url}
\usepackage{float}
\usepackage{stfloats}
\usepackage{algorithm}
\usepackage{algorithmic}
\usepackage{mathtools}

\usepackage{soul, color, xcolor}

\usepackage{colortbl}
\usepackage{caption}
\usepackage{eccv}

\usepackage{eccvabbrv}

\usepackage{graphicx}
\usepackage{booktabs}

\usepackage[accsupp]{axessibility}  

\usepackage{hyperref}

\usepackage{orcidlink}

\usepackage{pifont,xcolor}
\newcommand{\cmark}{\textcolor[rgb]{0,0.6,0}{\ding{51}}} 
\newcommand{\xmark}{\textcolor[rgb]{0.8,0,0}{\ding{55}}} 

\definecolor{bestcolor}{HTML}{E3EAF8}
\newcommand{\best}[1]{\cellcolor{bestcolor}{#1}}

\begin{document}

\title{VERTIGO: Visual Preference Optimization for Cinematic Camera Trajectory Generation} 

\titlerunning{VERTIGO}

\author{Mengtian Li\inst{1,2} \and
Yuwei Lu\inst{1} \and
Feifei Li\inst{1} \and
Chenqi Gan\inst{1} \and
Zhifeng Xie\inst{1,2} \and \\
Xi Wang\inst{3}\thanks{Corresponding author.}} 

\authorrunning{M.~Li et al.}

\institute{Shanghai University \and
Shanghai Engineering Research Center of Motion Picture Special Effects \and
LIX, \'Ecole Polytechnique, CNRS, IPP\\
\textcolor{magenta}{http://vertigo.magic-lab.tech/}}

\maketitle

{%
\centering
\includegraphics[width=0.98\linewidth]{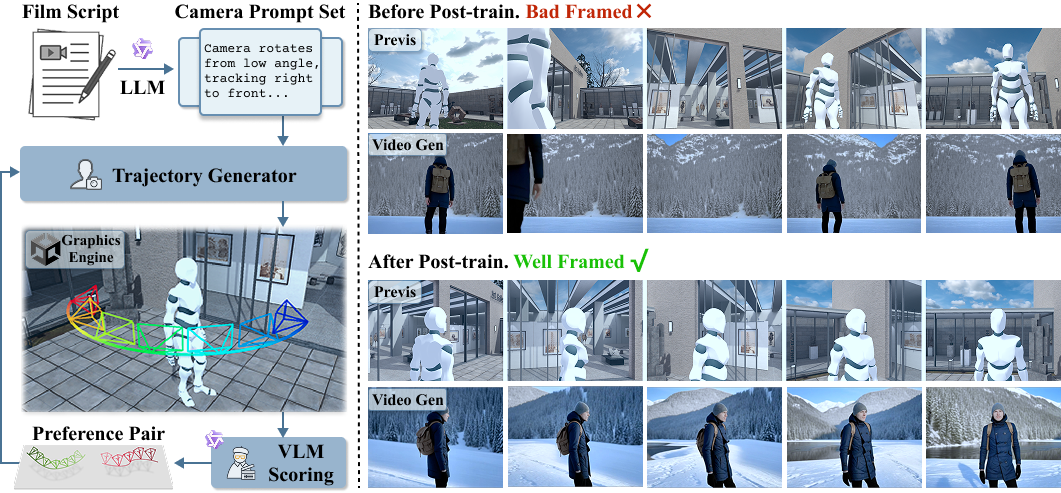}
\captionof{figure}{\textbf{Overview of VERTIGO.} We present an integrated framework that converts film scripts into 3D camera trajectories refined via preference-based post-training. On the right, we show GenDoP results before and after post-training, demonstrating improved framing and quality in both graphics engine rendering and video generation.}
\label{fig:teaser}
}

\begin{abstract}
Cinematic camera control relies on a tight feedback loop between director and cinematographer, where camera motion and framing are continuously reviewed and refined. Recent generative camera systems can produce diverse, text-conditioned trajectories, but they lack this ``director in the loop'' and have no explicit supervision of whether a shot is visually desirable. This results in in-distribution camera motion but poor framing, off-screen characters, and undesirable visual aesthetics.
In this paper, we introduce \textsc{VERTIGO}, the first framework for visual preference optimization of camera trajectory generators. Our framework leverages a real-time graphics engine (Unity) to render 2D visual previews from generated camera motion. A cinematically fine-tuned vision–language model then scores these previews using our proposed cyclic semantic similarity mechanism, which aligns renders with text prompts. This process provides the visual preference signals for Direct Preference Optimization (DPO) post-training.
Both quantitative evaluations and user studies on Unity renders and diffusion-based Camera-to-Video pipelines show consistent gains in condition adherence, framing quality, and perceptual realism. Notably, \textsc{VERTIGO} reduces the character off-screen rate from 38\% to nearly 0\% while preserving the geometric fidelity of camera motion. User study participants further prefer \textsc{VERTIGO} over baselines across composition, consistency, prompt adherence, and aesthetic quality, confirming the perceptual benefits of our visual preference post-training.

\end{abstract}

\section{Introduction}
\label{sec:intro}

In the film industry, cinematography plays a pivotal role in transforming textual scripts into visual narratives through camera movement, framing, and composition, conveying emotional and aesthetic meaning to the audience on the silverscreen. Traditional filmmaking relies on close collaboration between the cinematographer, who interprets the director’s instructions through camera movement, and the director, who supervises the on-screen composition from the monitor to decide whether a shot is kept or discarded. 

With recent advances in computer graphics and computer vision, particularly in generative AI, many works in computational cinematography have begun exploring automated or assisted camera control systems to emulate the creativity and precision of human cinematographers. 
Among recent developments, generative camera systems have shown great potential~\cite{jiang2024cinematographic,courant2024etexceptionaltrajectoriestexttocameratrajectory,zhang2025gendopautoregressivecameratrajectory}. Most are capable of producing diverse and dynamic camera trajectories conditioned on textual prompts. However, unlike real-world collaboration between the cinematographer and the film director, these systems lack a “\textit{director behind the monitor to validate or veto a shot}”, \emph{i.e.}, an \textbf{{informative feedback mechanism}} that guides the cinematographer in assessing whether the generated trajectories are aesthetically and narratively appropriate. In other words, simply generating in-distribution camera motions does not guarantee that the resulting shots exhibit good framing, spatial composition, or even the desired look-and-feel, a limitation also observed in recent work~\cite{courant2025pulp}.

This collaborative generation paradigm naturally points to post-training techniques that incorporate reward or preference signals~\cite{tong2025delving,huang2025patchdpo,liu2025improving,liu2025videodpo,zhu2025soft}. Such methods refine a pretrained generator by aligning it with downstream objectives through externally provided feedback, typically obtained from a reward model or a preference classifier that evaluates generated outputs. In our setting, this corresponds to equipping the camera generator with an additional evaluative signal capable of indicating whether a trajectory is desirable. 

However, unlike most visual generation tasks, camera trajectories cannot be \textit{directly} integrated into standard reward-based post-training frameworks. \textbf{Two main challenges arise}:  
(1) Unlike images or videos, a trajectory alone cannot be \textit{directly} interpreted visually, whereas, in real filmmaking, directors make decisions by watching the monitor, not by inspecting raw camera paths; and  (2) designing a reliable evaluator for such a domain-specific task requires constructing robust metrics and supervision signals that can faithfully assess cinematic quality.

In this paper, we propose leveraging a real-time graphics engine (e.g., Unity) to generate \textit{previews} (i.e., shot sequences rendered in the engine), which are then evaluated by a cinematically fine-tuned VLM using our cyclic semantic scoring mechanism that compares the semantic similarity between the original prompt and VLM-generated captions from rendered previews in embedding space. In this setup, the graphics engine functions as a 3D shooting stage, the camera generator plays the role of the cinematographer, and the VLM-based evaluator serves as the \textit{Director}. We then employ Direct Preference Optimization (DPO)~\cite{zhang2024directpreferenceoptimizationvideo} to post-train the camera trajectory generator on preference-ranked Unity previews, enabling the generator to better adhere to textual conditions, achieve more coherent framing, and produce trajectories with improved aesthetic look-and-feel.

Our method, \textsc{VERTIGO}, post-trains a GenDoP-style~\cite{zhang2025gendopautoregressivecameratrajectory} camera generator on our \textit{LenScript} dataset captioned with cinematic properties. Empirically, we observe consistent improvements in camera quality, particularly in condition adherence. To verify the visual improvement, we further evaluate \textsc{VERTIGO} using Unity rendering and diffusion-based Camera-to-Video generators, and report measurable gains in both rendered and generated video content. Most importantly, \textsc{VERTIGO} reduces the character off-screen rate from 38\% to nearly 0\%, while preserving, and in some cases even improving, geometric fidelity. User studies further confirm that \textsc{VERTIGO} achieves the highest preference rates in composition, consistency, prompt adherence, and aesthetic quality for both Unity-rendered previews and generated videos.
%


Our contributions are summarized as follows:
\begin{itemize}
    \item We are the first to explore a complete \textbf{visual reward–based post-training framework} for camera trajectory generation. Our system leverages real-time rendered frames as visual rewards and employs an effective and robust VLM-based cyclic semantic scoring mechanism to produce preference signals for reinforcement-style post-training.
    
    \item {We establish a reward system through our VLM-based cyclic semantic scoring mechanism, which computes latent semantic similarity between rendered shots and textual intentions. We proposed and compared multiple design choices and identified an effective strategy for using a VLM to evaluate trajectory quality from visual rewards, greatly improving the quality of framing.}

    \item 
    Through quantitative experiments and user evaluations, our DPO-trained camera generator produces improved trajectories in both Unity-rendered previews and camera-controlled video generation, demonstrating the effectiveness of the proposed reward and post-training framework.
\end{itemize}

\section{Related Work}

\subsection{Automatic Virtual Cinematography}

Recent advancements in virtual production, powered by real-time game engines\cite{unity, unreal}, have revolutionized filmmaking through dynamic previsualization. Early computational approaches established foundational frameworks via geometric and rule-based camera control~\cite{lino2015intuitive, christie2008camera, hayashi2014t2v, subramonyam2018taketoons, evin2022cine}, while recent systems leverage predefined cinematic concepts~\cite{rao2023dynamic, chen2024cinepregen, wei2025cinevisioninteractiveprevisualizationstoryboard} or LLM-driven multi-agent collaboration~\cite{xu2024filmagent}. Optimization-based methods manage camera parameters and integrate aesthetic principles~\cite{karakostas2020shot, louarn2018automated, galvane2015automatic, pueyo2024cinempc, bonatti2020autonomous}. Learning-based approaches employ reinforcement learning~\cite{gschwindt2019can, yu2022enabling}, deep networks~\cite{xie2023camera, yu2023automated}, and reference transfer~\cite{huang2019learning, huang2021one, wang2023jaws, jiang2024cinematic, jiang2021camera} to enhance trajectory quality. Emerging diffusion-based methods~\cite{jiang2024cinematographic, courant2024etexceptionaltrajectoriestexttocameratrajectory} generate plausible trajectories but struggle with multi-shot coherence and fine-grained control over shot scale, angle, and multi-segment planning. Current methods fail to holistically integrate narrative context and cinematographic principles with visual aesthetic quality.


\subsection{Controllable Video Generation}

Controllable camera motion is essential for film-oriented video generation. While recent methods enable 2D-based motion control ~\cite{guo2023animatediff, yang2024direct}, they lack explicit 3D spatial modeling, causing perspective inconsistencies. 3D-aware approaches like MotionCtrl ~\cite{wang2024motionctrl}, CameraCtrl ~\cite{he2024cameractrl}, and CVD ~\cite{kuang2024collaborative} condition on simplified trajectories yet omit intrinsic parameters and cinematographic principles. AKiRa ~\cite{wang2024akira} models intrinsics but requires nontrivial trajectory acquisition. GenDoP ~\cite{zhang2025gendopautoregressivecameratrajectory} comes closest by autoregressively generating camera poses, yet existing frameworks universally neglect visual aesthetics and fail to unify text-driven narrative intent, cinematographic trajectory planning, and video generation under a cohesive aesthetic-aware system. More comprehensive discussions of related work can be found in the supplementary material.

\subsection{Reward-based Fine-tuning with VLMs}
Reinforcement-learning-based post-training has become a prominent paradigm for aligning generative models with human preferences. In image and video generation, methods such as RLHF, RLAIF, and DPO have improved motion fidelity and text–visual consistency by distilling human-like preferences into diffusion models~\cite{ahn2024tuninglargemultimodalmodels,zhang2024directpreferenceoptimizationvideo,liu2024videodpoomnipreferencealignmentvideo}. Works like Control-A-Video~\cite{chen2024controlavideocontrollabletexttovideodiffusion} further demonstrate reward feedback learning for controllable video diffusion, using visual reward signals derived from rendered frames. Self-rewarding and AI-generated preference signals further reduce dependence on costly labels and help mitigate reward hacking~\cite{li2025selfrewardingvisionlanguagemodelreasoning,prabhudesai2024videodiffusionalignmentreward,liang2024richhumanfeedbacktexttoimage}. Recent Vision–Language Models (VLMs) also advance video understanding with finer temporal reasoning and narrative coherence~\cite{maaz2024videochatgptdetailedvideounderstanding,wang2024qwen2vlenhancingvisionlanguagemodels}, and benchmark suites such as ShotBench and VEU-Bench~\cite{liu2025shotbenchexpertlevelcinematicunderstanding,li2025veubenchcomprehensiveunderstandingvideo} probe cinematographic perception including shot scale, camera movement, and editing style. However, unlike images or videos where pixel-level visual feedback is readily available, camera trajectories are purely geometric sequences that carry no visual content on their own, making it impossible to directly derive visual reward signals. Applying VLM-derived rewards to camera-trajectory generation thus remains largely unexplored, requiring mechanisms that bridge trajectory space and visual space to enable preference-based optimization.
\section{Method}
Unlike prior approaches that focus primarily on geometric aspects of camera generation, our goal is to use fast, geometrically consistent rendering previews from a graphics engine (e.g., Unity) as \textit{visual reward signals} to post-train camera generators in a reinforcement learning paradigm, so that the resulting camera motions lead to well-framed, aesthetically pleasing shots in both rendering-based and generation-based video production pipelines.

\begin{figure}[h]
  \centering
  \includegraphics[width=\columnwidth]{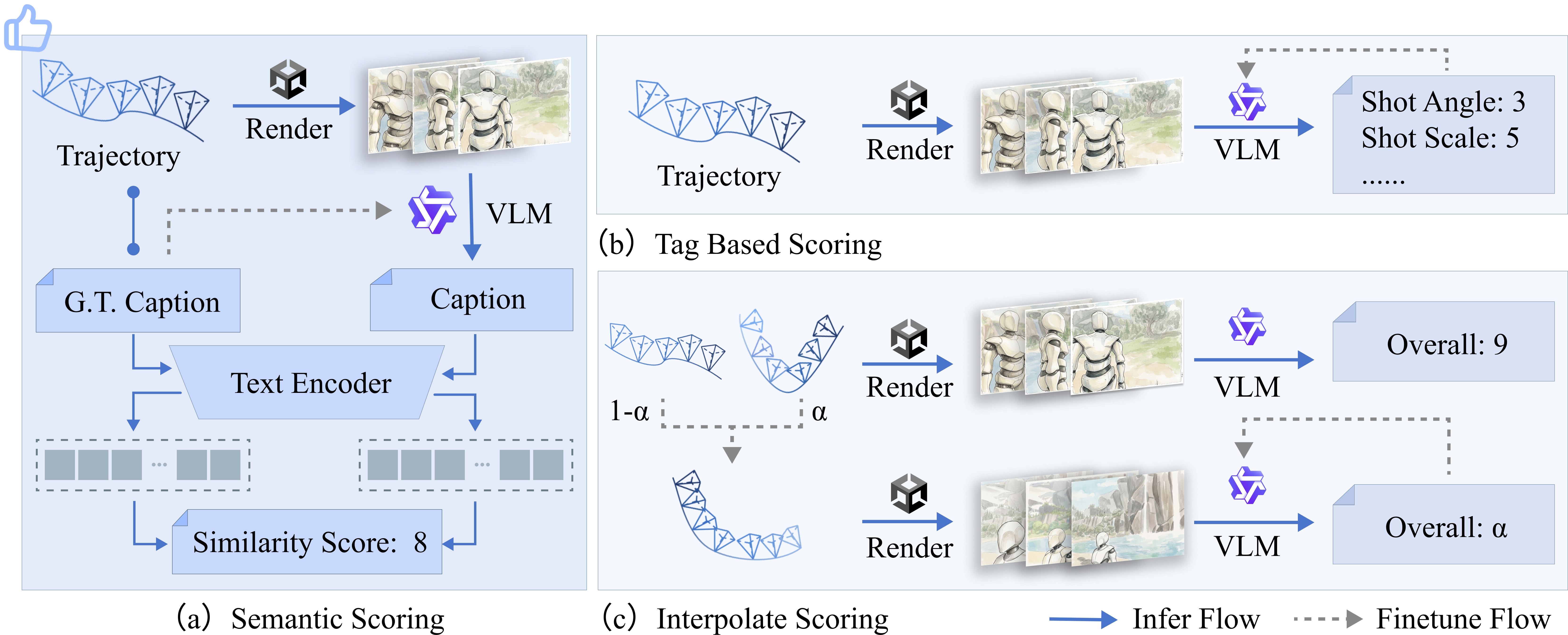}
   \caption{\textbf{Different VLM scoring and fine-tuning strategies of VERTIGO.} We explore three preference scoring methods: (a) cyclic semantic scoring via latent-space similarity. (b) tag-consistency scoring; (c) direct scalar regression via RAFT-style fine-tuning on interpolated trajectories;}
   \label{fig:onecol}
   \vspace{-2mm}
\end{figure}

Central to our approach is a \textit{visual feedback construction mechanism} that bridges the gap between trajectory-only camera generators and visual-quality assessment. Since raw trajectories carry no visual content, conventional post-training methods cannot directly evaluate or optimize their cinematic quality. Our mechanism leverages real-time graphics-engine rendering to materialize trajectories into preview sequences, and employs a VLM-based cyclic semantic scorer to derive preference signals, forming an automated, \textit{render-in-the-loop} post-training pipeline. Concretely, the pipeline consists of three core technical components:
(1) utilizing a pre-constructed Unity environment to render views generated by a pretrained camera generator;
(2) employing cyclic semantic scoring within a VLM fine-tuned for camera motion classification to evaluate generated and augmented trajectories;
(3) applying a DPO-style objective with positive and negative examples to post-train the camera generator. The complete pipeline is illustrated in Figure~\ref{fig:onecol}, with detailed descriptions of each component provided in subsequent sections.




\begin{figure*}[!t]
  \centering
  \includegraphics[width=\textwidth,]{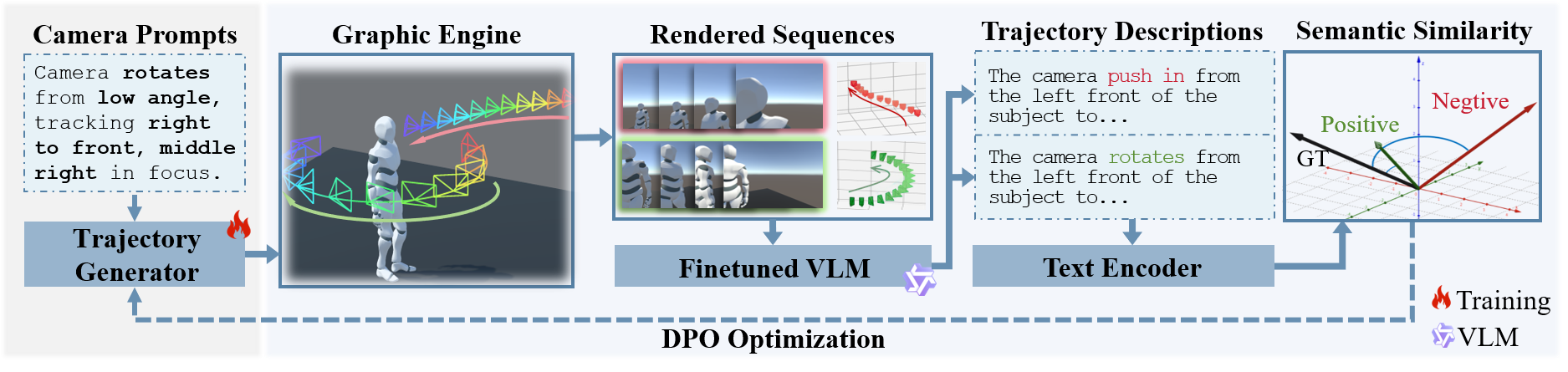}
  \caption{\textbf{Pipeline of VERTIGO.} From a camera prompt, the generator produces 3D trajectories rendered into preview sequences by a graphics engine. A VLM performs inverse reasoning to caption the realized motion; the original prompt and generated caption are compared in latent space to derive preference scores for DPO post-training.
}
  \label{Method}
  \vspace{-2mm}
\end{figure*}

\subsection{Trajectory Generator}

We formalize the camera trajectory generation task as a conditional generation problem. Given a natural-language prompt $p$ describing cinematographic intent and an optional conditioning context $c$ (e.g., scene layout, previous keyframe), the generator $\pi_\theta$ samples a camera trajectory $\tau$ from the learned distribution $\pi_\theta(\tau \mid p, c)$, parameterized by $\theta$. During inference, the model produces temporally coherent camera sequences that respect both geometric constraints and narrative semantics encoded in $p$. This abstraction is compatible with various generation paradigms (e.g., diffusion or auto-regressive models) and serves as the basis for subsequent preference-based post-training.


\noindent
\textbf{Dataset: LenScript.} Existing datasets~\cite{courant2024etexceptionaltrajectoriestexttocameratrajectory, zhang2025gendopautoregressivecameratrajectory} reconstructed from videos lack fine-grained control over framing, shot composition, and focal properties, limiting their utility for training models that must reason about on-screen appearance. Since our trajectories are intended for previsualization and visual evaluation, we require training data that explicitly encodes cinematic intent at both geometric and visual levels. To this end, we procedurally construct \textbf{LenScript}, a trajectory dataset synthesized in Unity with precise annotations for both camera motion patterns and compositional attributes. Following the categorical structure of CCD~\cite{jiang2024cinematographic}, each trajectory is tagged with cinematic dimensions and paired with natural-language captions. Trajectories are synthesized in a local coordinate system and serialized in RealEstate10k-compatible format, supporting continuous focal length control. This dataset serves as the foundation for pretraining our trajectory generator, fine-tuning the VLM evaluator, and conducting ablation studies. Detailed dataset statistics are provided in the supplementary material.

\noindent
\textbf{Graphics Engine.} Once the generator $\pi_\theta$ is trained, we deploy it in a Unity environment with multiple pre-constructed scenes for trajectory preview, editing, and render-sequence export. Its trajectories are rendered into visual previews and passed to the VLM for preference scoring and post-training supervision. This engine-native pipeline ensures geometrically consistent, visually grounded evaluation of camera motion across diverse scene configurations.

\subsection{Trajectory Preference in VLM}
\label{sec:vlm-preference}


\textbf{Motivation.} To post-train the trajectory generator, we require clear and precise supervision signals. However, evaluating camera trajectories and their rendered previews is inherently challenging and multi-faceted, as factors such as framing, motion rhythm, and narrative intent rarely correlate with geometric or pixel-level errors. Unlike handcrafted heuristics that struggle with compositional nuances, vision-language models naturally ground visual cinematography in semantic space and enable holistic quality assessment through learned aesthetic representations.


More specifically, our preference scoring process proceeds as follows: 
given a textual prompt $p$, the trajectory generator samples a set of candidate camera motions 
$\{\tau_i\}_{i=1}^{N}$ using temperature and top-$k$ sampling to promote exploration. 
Each trajectory $\tau_i$ is then rendered by the engine into a sequence of frames 
$\mathcal{I}_i = \{I_i^t\}_{t=1}^{T}$, along with a corresponding 3D path visualization. 
The goal is to evaluate how well each rendered preview $\mathcal{I}_i$ aligns with the textual intent expressed in $p$.  We discuss three different approaches to achieve this:


\noindent\textbf{(1) Tag-consistency scoring.} 
We consider five canonical cinematography dimensions—the same categories as CCD’s synthesis pipeline ~\cite{jiang2024cinematographic}: Camera Movement, Shot Scale, Shot Direction, Shot Angle, and Screen Property. For each dimension specified or implied by the prompt, the VLM judges whether the rendered sequence satisfies it, yielding a count of matched dimensions in the range 0 to 5. We then discretize this count into a 10-level integer score $s_i^{\text{tag}}\in\{0,\dots,9\}$ (higher is better), reflecting how many of the five dimensions are correctly satisfied. 

\noindent\textbf{(2) Direct scalar regression.} We prompt the VLM to reason about the five canonical dimensions and synthesize its judgments into a single aggregated match score $s_i^{\text{reg}}\in[0,9]$ that reflects framing quality, motion rhythm, and compositional balance between $\mathcal{I}_i$ and $p$. This treats preference estimation as a unified scalar judgment rather than per-dimension verification. To obtain graded supervision and reduce labeling sparsity, we synthesize interpolated trajectories $\tilde{\tau} = \alpha\,\tau_i + (1-\alpha)\,\tau_j$ ($\alpha\in[0,1]$) between paths from different prompts, render the corresponding sequence $\tilde{\mathcal{I}}$, and assign $\hat{y}=\operatorname{round}(9 \times \alpha)$ as the continuous supervision target. We then minimize an auxiliary RAFT-style token regression loss~\cite{lukasik2024regression, lukasik2025better} that aligns the scalar digit with the auto-regressive token distribution:
\begin{equation}
\mathcal{L}_{\text{RAFT}} \,=\, \frac{1}{|\mathcal{D}|} \sum_{t\in\mathcal{T}_d} \Bigg( \hat{y}_t - \sum_{d\in\mathcal{D}} p_t(d)\, d \Bigg)^2,
\end{equation}
where $p_t(d)$ is the predicted token distribution at decoding step $t$, $\mathcal{D}$ denotes the discretized score vocabulary, and $\hat{y}_t$ repeats the target score at each step to enforce consistent output across the generation sequence.

\noindent\textbf{(3) Cyclic Semantic Scoring.} Inspired by cycle-consistency principles in reward learning ~\cite{bahng2025cycleconsistencyrewardlearning}, we let the VLM perform inverse cinematic reasoning, producing a natural-language description $\hat{p}_i$ that summarizes the camera behavior in $\mathcal{I}_i$. We then encode both the original prompt $p$ and the generated caption $\hat{p}_i$ using a dedicated semantic embedding model $\phi(\cdot)$ and compute their cosine similarity in the latent space:
\begin{equation}
s_i^{\text{sem}} = \frac{\phi(p) \cdot \phi(\hat{p}_i)}{\|\phi(p)\|\,\|\phi(\hat{p}_i)\|}.
\end{equation}

To enhance discriminative sensitivity on cinematographic captions, we apply lightweight LoRA contrastive fine-tuning on the embedding model using caption pairs from the LenScript dataset. This cyclic scoring mechanism measures semantic alignment between intent and realized camera motion, providing grounded preference signals for DPO.

Within each prompt-conditioned set, we apply one scoring method to rank trajectories and construct preference pairs $(\tau_i,\tau_j)$ where $s_i > s_j$.

\noindent
\textbf{Choice of scoring strategy.} Empirically, we find that both tag-consistency and direct regression suffer from inherent limitations when applied to cinematographic evaluation. VLMs struggle to provide fine-grained numerical judgments for camera trajectories, as discrete categorical scores (0–9) tend to cluster around a narrow range, losing discriminative power across visually similar but qualitatively different camera motions, which in turn hampers reliable construction of preference pairs. Moreover, these schemes exhibit weaker generalization across scenes and shot styles, and are brittle under prompt under specification. As a result, the regression variant tends to collapse to low-variance outputs (mode collapse), producing unreliable and poorly differentiated scores on Unity-rendered sequences with trajectory visualizations overlaid in 3D space. By contrast, cyclic semantic scoring provides richer signals through caption-space comparison, improving robustness to domain shift and enabling more diverse and stable preference ordering. Unless otherwise noted, we therefore adopt cyclic semantic scoring as the default scorer for all subsequent experiments.




\noindent
\textbf{VLM Fine-tuning for Cinematic Tasks.} Off-the-shelf VLMs are trained on natural images and struggle with the domain gap introduced by our Unity-based renders, trajectory visualizations, and stylized lighting. This mismatch often leads the evaluator to misinterpret motion cues or framing, resulting in unstable preference judgments. To mitigate this, we start from \textbf{ShotBench}~\cite{liu2025shotbenchexpertlevelcinematicunderstanding}, a VLM already fine-tuned on large-scale cinematography data, and further adapt it to the Unity-rendered domain using lightweight LoRA fine-tuning. The model is trained on pairs of rendered sequences and natural-language camera captions, requiring no additional synthetic labels beyond the captions themselves.

\subsection{Post-training for Trajectory Generation}\label{sec:post-train}


\textbf{Motivation.} Supervised imitation alone does not align generation with nuanced cinematic preferences, as it treats all training samples equally and tends to overfit to seen prompts without distinguishing relative quality among plausible trajectories. Preference-based post-training addresses this by directly optimizing pairwise quality comparisons, regularizing toward a reference policy, and improving cinematic realism without brittle reward engineering.

To this end, we employ Direct Preference Optimization (DPO) to post-train the trajectory generator with rendered-view preferences, using preference pairs automatically derived from the VLM-based scoring scheme. For each prompt $p$, candidate trajectories are ranked, and pairs $(\tau_i,\tau_j)$ are formed where $\tau_i$ is preferred to $\tau_j$. DPO optimizes a policy to increase the likelihood of preferred sequences relative to a frozen reference model, without the need to train an explicit reward model. Concretely, the relative reward is defined as $r_{\theta}(\tau \mid p) \coloneqq \log \frac{\pi_{\theta}(\tau \mid p)}{\pi_{\text{ref}}(\tau \mid p)}$, where $\pi_{\theta}$ is the learnable policy and $\pi_{\text{ref}}$ is a pretrained reference. The corresponding loss:
\begin{equation}
\mathcal{L}_{\text{DPO}} = -\mathbb{E}_{(\tau_i, \tau_j)} \!\left[
\log \sigma\!\left(
\beta \left( 
r_{\theta}(\tau_i \mid p) - r_{\theta}(\tau_j \mid p)
\right)
\right)
\right],
\end{equation}
encourages correct pairwise ordering while implicitly regularizing toward $\pi_{\text{ref}}$, thereby preserving diversity and mitigating over-optimization.

Intuitively, this objective increases the log-odds of preferred trajectories relative to a fixed reference policy, serving as an implicit regularizer that limits distributional drift while improving alignment with the preference signal. The scalar $\beta$ controls the pairwise margin, adjusting how strongly the policy diverges from its reference. In practice, optimizing this objective produces trajectories with higher preference scores: showing smoother motion, cleaner framing, and more stable spatial relations around targets.

\section{Experiment}

\begin{table*}[t]
  \centering
  \footnotesize
    \caption{\textbf{Quantitative comparison of camera trajectory and video quality.} We evaluate geometric fidelity using CLaTr-based metrics ~\cite{courant2024etexceptionaltrajectoriestexttocameratrajectory} and visual quality via VBench ~\cite{li2024mvbenchcomprehensivemultimodalvideo} metrics on both Unity-rendered sequences and controllable video generation. VERTIGO achieves competitive geometric performance while substantially improving visual metrics and target framing stability. \colorbox{bestcolor}{Shaded} indicates best performance.}
  \resizebox{\textwidth}{!}{%
  \begin{tabular}{p{7.78em}|cccccc|ccc|cc}
    \toprule
    \multicolumn{1}{c|}{\multirow{2}[2]{*}{\textbf{Methods}}} 
    & \multicolumn{6}{c|}{\textbf{Camera trajectory quality}} 
    & \multicolumn{3}{c|}{\textbf{Render Quality}} 
    & \multicolumn{2}{c}{\textbf{Video Quality}} \\
    
    \multicolumn{1}{c|}{} 
    & FCD↓ & CS↑ & P↑ & R↑ & D↑ & C↑ 
    & MisR.↓ & Cons.↑ & Aes.↑ 
    & Cons.↑ & Aes.↑ \\
    \hline
    
    CCD\hfill~\cite{jiang2024cinematographic}   
    & 104.60 & 30.03 & 0.58 & 0.01 & 0.41 & 0.10 
    & 0.999 & 0.820 & 0.460 
    & 0.901 & 0.296 \\
    
    DIRECTOR\hfill~\cite{courant2024etexceptionaltrajectoriestexttocameratrajectory}
    & 26.28 & 32.20 & 0.87 & 0.29 & 0.79 & 0.44 
    & 0.803 & 0.892 & 0.464 
    & 0.908 & 0.293 \\
    
    GenDoP\hfill~\cite{zhang2025gendopautoregressivecameratrajectory} 
    & \best{4.17} & 67.98 & \best{0.92} & 0.72 & \best{0.93} & \best{0.78} 
    & 0.387 & 0.904 & 0.515 
    & 0.967 & 0.722\\
    \hline
    VERTIGO 
    & 4.22 & \best{68.40} & \best{0.92} & \best{0.73} & \best{0.93} & \best{0.78} 
    & \best{0.008} & \best{0.908} & \best{0.518} 
    & \best{0.969} & \best{0.735} \\
    
    \bottomrule
  \end{tabular}%
  }%
  \label{tab:quantitative}%
\end{table*}

\begin{figure*}[t]
  \centering
  \includegraphics[width=\linewidth,scale=1.00]{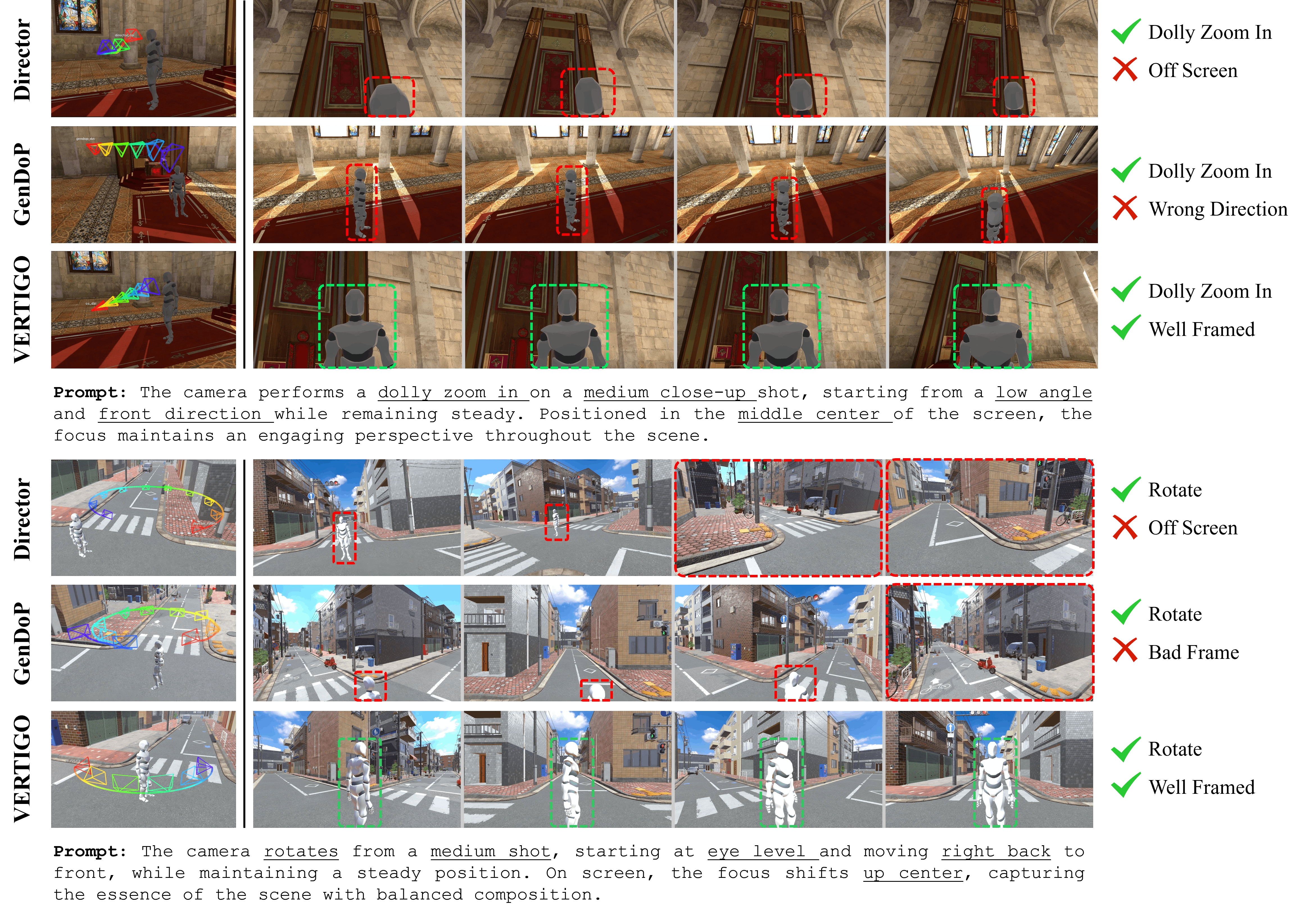}	
  \caption{\textbf{Qualitative comparison of camera generators.} VERTIGO accurately adheres to spatial composition instructions and maintains framing, whereas GenDoP and DIRECTOR misplace subjects and occasionally lose them.}	
  \label{qualitative}
\end{figure*}

\begin{figure*}[t]
  \centering
  \includegraphics[width=\linewidth,scale=1.00]{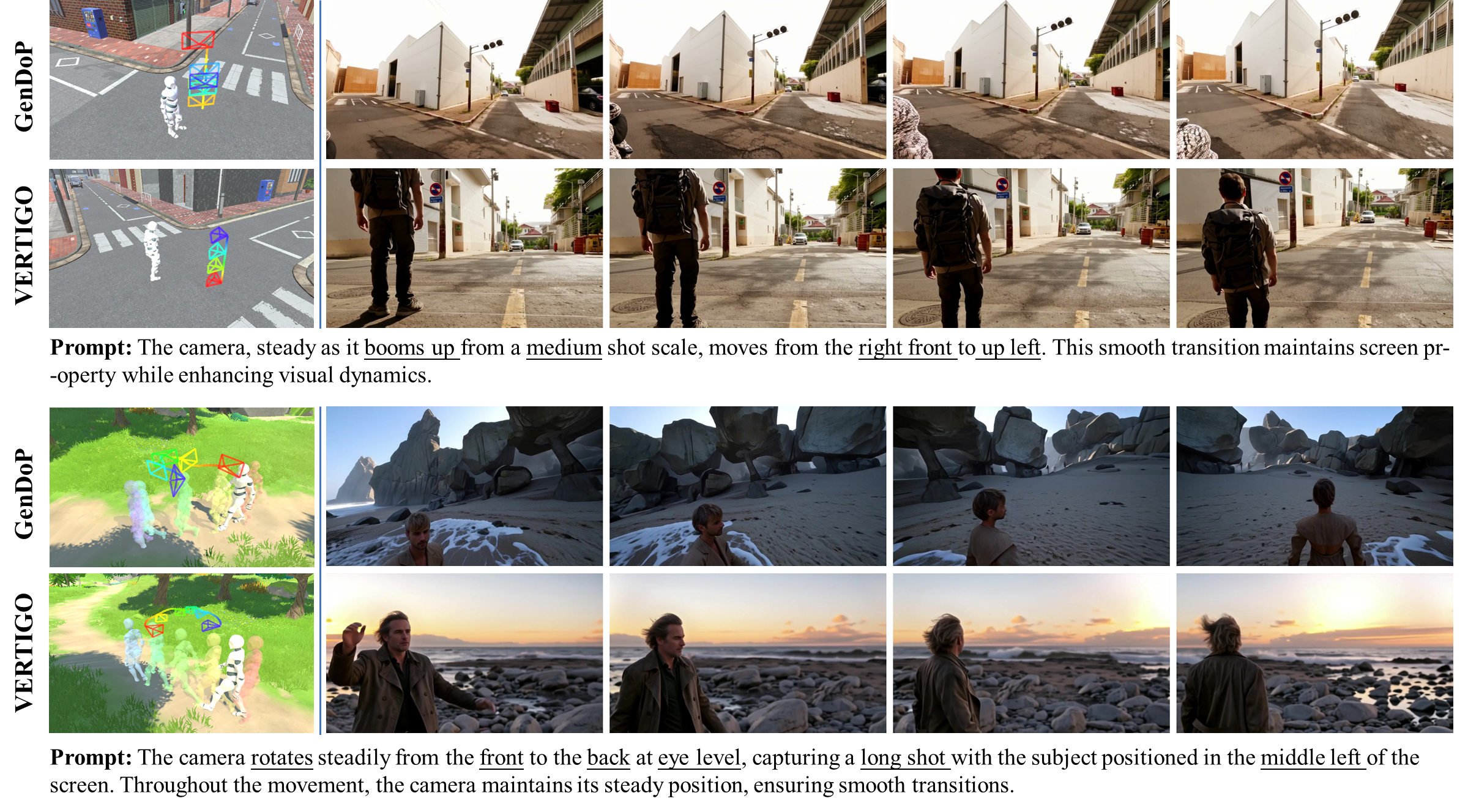}	
  \caption{\textbf{Qualitative comparison of video-to-video transfer results.} We compare VACE-based video transfer on two trajectories. For the first trajectory, we show GenDoP vs.\ VERTIGO transfer results; for the second, we compare static and animated character renderings transferred under the same trajectory. VERTIGO produces more robust framing after transfer, while character animation does not affect transfer quality, as trajectories are defined in the subject's local coordinate space.}	
  \label{qualitative_video}
\end{figure*}
\noindent

\noindent\textbf{Setup.} 
We evaluated the trajectory generation quality across geometric and visual dimensions. Geometric evaluation employed CLaTr-based~\cite{courant2024etexceptionaltrajectoriestexttocameratrajectory} metrics to assess all methods on the full test set. For visual evaluation, we uniformly sample 1,000 trajectories from the test set under two conditions and measure their quality via VBench metrics~\cite{huang2023vbenchcomprehensivebenchmarksuite}:
(1) Direct rendering in four Unity scenes for visual content;
(2) Video generation by transforming Unity-rendered videos using Wan 2.1 VACE~\cite{vace}, ensuring a comprehensive assessment of both geometric fidelity and downstream visual impact across rendering and generation paradigms.

\noindent\textbf{Configs.}
We employ the Qwen 2.5‑VL~\cite{wang2024qwen2vlenhancingvisionlanguagemodels}, fine‑tuned on ShotBench~\cite{liu2025shotbenchexpertlevelcinematicunderstanding}, as the VLM for our preference scoring framework. For cyclic semantic scoring, we use a domain-adapted E5~\cite{wang2024multilinguale5textembeddings} text embedding model, fine-tuned on cinematographic captions from our LenScript dataset. 
All experiments are conducted on two NVIDIA RTX 5880 Ada GPUs.
Notably, our visual feedback construction pipeline—including real-time engine rendering, VLM-based preference scoring, and DPO optimization—is computationally lightweight. The graphics engine provides rendering that is orders of magnitude faster than diffusion-based video generators, and a compact VLM is sufficient for reliable cinematographic evaluation, enabling the entire post-training loop to converge efficiently without imposing significant computational overhead.

\noindent\textbf{Metrics.}
 We evaluate our method using two complementary groups of metrics measuring geometry-level trajectory quality and video-level perceptual quality. 
 \begin{itemize}
\item\textit{Geometrical Metrics}. At the geometry level, we assess camera trajectory quality using a series of CLaTr-based metrics~\cite{courant2024etexceptionaltrajectoriestexttocameratrajectory}, including the Frechét CLaTr Distance (FCD), CLaTr-Score (CS), Precision (P), Recall (R), Density (D), and Coverage (C), which collectively quantify the fidelity and diversity of the generated trajectories with respect to the ground-truth distribution. To ensure a fair comparison, we retrain the CLaTr baseline on both our Lenscript and the E.T. dataset.
\item\textit{Visual Metrics}. At the video level, we further evaluate the perceptual impact of different trajectory generation models when applied to video synthesis. Following the VBench~\cite{huang2023vbenchcomprehensivebenchmarksuite}, we measure consistency, and aesthetic, which respectively capture the temporal dynamic coherence, motion continuity, and perceptual appeal of the rendered video sequences. We also evaluate the Missing Rate (MisR) of the shooting target in Unity, which measures the fraction of frames where the shooting target is either off-screen or positioned within the outer 20\% border region to quantify the framing failure of the generated trajectory. Together, these metrics provide a comprehensive evaluation of both the structural realism of camera trajectories and their downstream effect on video generation quality.
\end{itemize}

\subsection{Quantitative Results}

VERTIGO demonstrates strong performance across both geometric and visual quality metrics. As shown in Table~\ref{tab:quantitative}, we achieve competitive geometric metrics comparable to the current state-of-the-art trajectory generator in terms of competing FCD and enhanced CLaTr-Score, demonstrating that DPO post-training preserves distributional alignment with ground-truth camera motions.

Critically, while preserving geometric fidelity, our method substantially outperforms all baselines on visual quality metrics in both Unity-rendered and generated video settings. For example, \textsc{VERTIGO} reduces the Missing Rate (MisR) to $0.008$, compared to $0.387$ for GenDoP, $0.803$ for DIRECTOR, and $0.999$ for CCD. In other words, our post-training nearly eliminates target framing failures. This directly reflects \textbf{the efficacy of visual-aware post-training} in preventing off-screen or edge-biased target placements that frequently occur in geometry-only approaches, and validates our core hypothesis that \textbf{rendered preview feedback is essential for stable target framing}. Beyond framing stability, our method also yields consistent gains on video-based metrics from VBench~\cite{li2024mvbenchcomprehensivemultimodalvideo}. 
In the Unity rendering setting, \textsc{VERTIGO} improves both Consistency (0.908) and Aesthetic Quality (0.518), and in the video generation setting, it further achieves 0.969 Consistency and 0.735 Aesthetic Quality. These results show that our preference post-training with rendered previews successfully bridges geometric correctness and visual composition, producing trajectories that are structurally sound, visually compelling, and ready for downstream cinematic use in both CG and generative pipelines.

\subsection{Qualitative Results}

We present qualitative comparisons of trajectories generated by GenDoP, DIRECTOR and our method, rendered directly in Unity under identical prompts.

\noindent\textbf{Prompt Adherence}. Both GenDoP and DIRECTOR demonstrate reasonable compliance with basic motion instructions while exhibiting failures in handling compositional aspects that require spatial reasoning. For instance, when the user prompt specifies ``positioned in the middle center of the screen'', both baselines fail to correctly interpret this spatial constraint, misplacing the target. In contrast, VERTIGO accurately grounds the compositional instruction, precisely localizing the target within the specified screen region.

\noindent\textbf{Visual Quality.} Existing models often struggle to maintain focus on the target, sometimes even allowing it to disappear from the frame. This shortcoming stems from their geometric-only training paradigm, which neglects 2D visual feedback from the rendered image. VERTIGO overcomes this limitation through the proposed visual feedback construction mechanism, which enables render-in-the-loop post-training by materializing trajectories into preview sequences for VLM-based preference evaluation. As shown in Figure~\ref{qualitative}, our method executes the correct motion pattern and faithfully preserves the ``medium shot'' composition, ensuring the subject remains properly framed as dictated by the prompt. We further evaluate video-to-video transfer quality by transforming Unity-rendered sequences using VACE under identical prompts. As shown in Figure~\ref{qualitative_video}, we compare GenDoP and VERTIGO transfer results on one trajectory, and additionally compare static versus animated character transfers on a second trajectory. VERTIGO consistently maintains superior framing after transfer, while character animation has no effect on transfer quality, since trajectories are defined in the subject's local coordinate space. These results demonstrate that our visual preference signals transfer robustly to downstream video synthesis and generalize naturally across both static and dynamic subjects.

\noindent\textbf{Animated Characters.} We further evaluate our method on scenes with animated, non-static characters to verify compatibility with dynamic subjects. Since our trajectory is defined in a local coordinate system relative to the subject, camera poses are applied frame-by-frame to the subject's current position, naturally accommodating character animation. As illustrated in Figure~\ref{qualitative_video}, VERTIGO maintains consistent framing and composition quality on animated characters, demonstrating that the visual feedback mechanism captures motion-agnostic framing principles rather than overfitting to static-scene configurations. Moreover, since our post-training operates solely on camera trajectories and is agnostic to rendering style, the learned preferences generalize naturally to diverse visual styles, from photorealistic rendering to stylized and artistic domains. Additional qualitative comparisons are provided in the supplementary materials.

\subsection{Ablation Study}

To validate the design choice and efficacy of our cyclic semantic scoring strategy in Sec.\,\ref{sec:vlm-preference}, we conduct an ablation study comparing all three candidate scoring methods. We post-train the camera generator with preference pairs constructed from different scoring methods, then evaluate the results on Unity-rendered sequences using VBench metrics.

\begin{table*}[t]
  \centering
  \footnotesize
  \vspace{-6pt}
  \setlength{\tabcolsep}{4pt}
    \caption{\textbf{Ablation study.} Starting from the GenDoP baseline, we compare three VLM-based preference scoring strategies for VERTIGO, evaluate generalization on animated characters, and ablate DPO $\beta$. All models are evaluated on Unity-rendered sequences using VBench metrics.}
  \resizebox{\textwidth}{!}{%
  \begin{tabular}{@{}l|c|ccc|ccc|cc@{}}
    \toprule
    \multirow{2}{*}{Metric}
    & \multirow{2}{*}{GenDoP}
    & \multicolumn{3}{c|}{VERTIGO (Scoring Strategy)}
    & \multicolumn{3}{c|}{VERTIGO ($\beta$ ablation)}
    & \multicolumn{2}{c@{}}{Anim.\ Characters} \\
    &
    & Tag & Interp. & Sem. ($\beta{=}0.1$)
    & $\beta{=}0.01$ & $\beta{=}0.5$ & $\beta{=}0.9$
    & GenDoP & Ours \\
    \hline
    MisR.$\downarrow$
    & 0.387
    & 0.327 & 0.286 & \best{0.008}
    & 0.097 & 0.132 & 0.414
    & 0.387 & 0.008 \\
    Cons.$\uparrow$
    & 0.904
    & 0.902 & 0.906 & \best{0.908}
    & 0.897 & 0.905 & 0.901
    & 0.876 & 0.902 \\
    Aes.$\uparrow$
    & 0.515
    & 0.516 & 0.515 & \best{0.518}
    & 0.482 & 0.517 & 0.515
    & 0.479 & 0.488 \\
    \bottomrule
  \end{tabular}%
  }%
  \label{tab:ablation}
  \label{tab:combined}
\end{table*}

\noindent\textbf{Scoring Strategy.} Our cyclic semantic scoring outperforms other strategies while avoiding mode collapse. As shown in Table~\ref{tab:ablation}, both tag-consistency and scalar regression exhibit significant limitations during DPO training. Discrete numerical scores lack sufficient granularity to distinguish subtle cinematographic variations, producing low-variance outputs that hamper reliable construction of preference pairs and fail to provide effective guidance for optimization. In contrast, cyclic semantic scoring leverages the rich semantic space of natural language captions, preserving fine-grained distinctions in camera behavior and framing intent. Our method achieves the best performance across all metrics, with dramatically reduced Missing Rate and improved consistency and aesthetic quality, validating that semantically grounded preference learning is essential for high-quality trajectory generation.

\noindent\textbf{DPO Temperature $\beta$.} We ablate the temperature parameter $\beta$, which controls how strongly the policy may diverge from the reference model. As shown in Table~\ref{tab:ablation}, $\beta{=}0.1$ yields the best overall trade-off, achieving the lowest Missing Rate while maintaining high consistency and aesthetic quality. Smaller values (e.g., $\beta{=}0.01$) under-regularize the policy, leading to degraded framing stability, while larger values ($\beta \geq 0.5$) over-constrain optimization and limit the model's capacity to learn effective visual preferences from the rendered feedback.

\noindent\textbf{Animated Characters.} To evaluate generalization beyond static scenes, we further test VERTIGO on sequences featuring animated characters with dynamic motion. As shown in Table~\ref{tab:ablation}, our method consistently improves framing quality over GenDoP on animated subjects, reducing the Missing Rate while improving both consistency and aesthetic scores. These results confirm that the visual preference signals acquired through our render-in-the-loop post-training transfer effectively to dynamic character scenarios, demonstrating that the learned framing behaviors are not tied to static-scene configurations.






\subsection{User Study}

To assess the perceived cinematic quality of generated trajectories, we conducted a human evaluation comparing VERTIGO against CCD, DIRECTOR, and GenDoP. We recruited 34 participants (11 experts with cinematography experience and 23 general participants) across two settings: (1) \textit{Unity rendering}, directly reflecting trajectory quality, and (2) \textit{video-to-video transfer} via Wan 2.2 VACE~\cite{vace}, assessing stylization fidelity and transfer resilience. Within each group of four videos (one per method, presented in randomized order), participants performed a \textit{best-of-4 selection} task, choosing the best video along five perceptual dimensions per setting. Full questionnaire design and additional results (including full ranking and further comparisons) are provided in the supplementary materials.

\begin{figure}[t]
  \centering
  \includegraphics[width=\columnwidth]{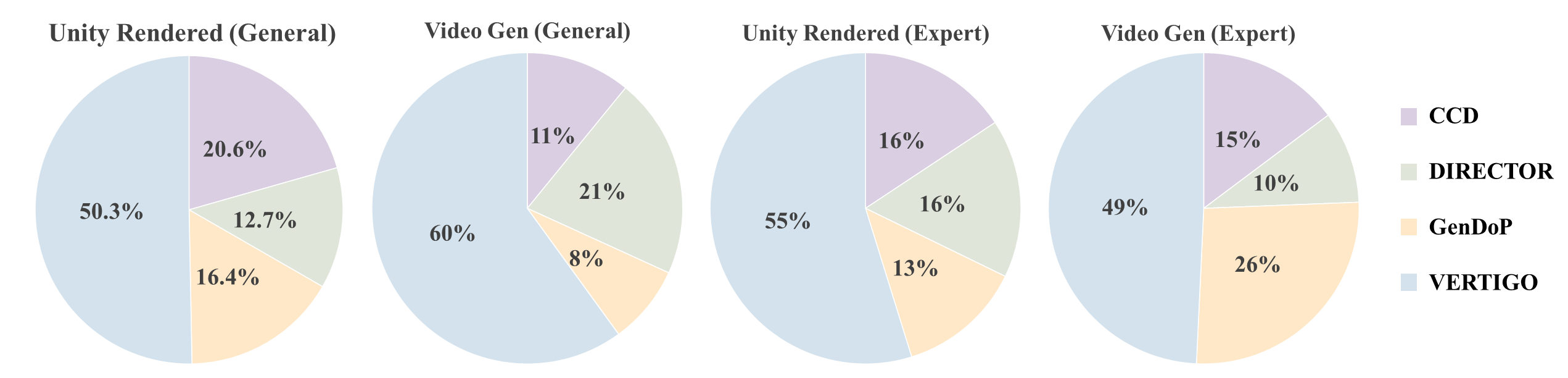}
  \caption{\textbf{User study best-of-4 results.} Preference rates across evaluation dimensions for Unity rendering and video-to-video transfer (Wan 2.2 VACE). See supplementary for questionnaire details and additional results.}
  \label{fig:user_study}
\end{figure}

As shown in Figure~\ref{fig:user_study}, VERTIGO achieves the highest preference rates across all evaluation dimensions in both settings, with an overall win rate of 52.4\%, excelling in composition (51.8\%) and instruction adherence (53.9\%). These results confirm that our visual preference post-training effectively improves perceptual trajectory quality across both CG rendering and generative transfer modalities.

\section{Conclusion}

We present \textsc{VERTIGO}, a visual preference optimization framework that post-trains camera-trajectory generators using computer-graphics–rendered previews as visual reward signals. To tackle the challenging problem of evaluating trajectories through their resulting visual content, we introduce a VLM-based cyclic semantic scoring mechanism for assessing cinematographic quality, and we systematically compare it against multiple alternative scoring designs through extensive ablations. Experiments show that our approach (1) dramatically reduces bad framings and improves prompt adherence while preserving high geometric fidelity, and (2) consistently enhances visual quality across both rendering-based and diffusion-based video generation pipelines. These improvements are corroborated by quantitative metrics and user studies. As the first framework to explicitly bridge geometric camera generation with the shot visual outcomes, \textsc{VERTIGO} pushes computational cinematography toward practical, director-in-the-loop visual generation and broader real-world creative applications.


%
%

\begin{thebibliography}{10}
\providecommand{\url}[1]{\texttt{#1}}
\providecommand{\urlprefix}{URL }
\providecommand{\doi}[1]{https://doi.org/#1}

\bibitem{ahn2024tuninglargemultimodalmodels}
Ahn, D., Choi, Y., Yu, Y., Kang, D., Choi, J.: Tuning large multimodal models for videos using reinforcement learning from ai feedback (2024), \url{https://arxiv.org/abs/2402.03746}

\bibitem{bahng2025cycleconsistencyrewardlearning}
Bahng, H., Chan, C., Durand, F., Isola, P.: Cycle consistency as reward: Learning image-text alignment without human preferences (2025), \url{https://arxiv.org/abs/2506.02095}

\bibitem{bonatti2020autonomous}
Bonatti, R., Wang, W., Ho, C., Ahuja, A., Gschwindt, M., Camci, E., Kayacan, E., Choudhury, S., Scherer, S.: Autonomous aerial cinematography in unstructured environments with learned artistic decision-making. Journal of Field Robotics  \textbf{37}(4),  606--641 (2020)

\bibitem{chen2024controlavideocontrollabletexttovideodiffusion}
Chen, W., Ji, Y., Wu, J., Wu, H., Xie, P., Li, J., Xia, X., Xiao, X., Lin, L.: Control-a-video: Controllable text-to-video diffusion models with motion prior and reward feedback learning (2024), \url{https://arxiv.org/abs/2305.13840}

\bibitem{chen2024cinepregen}
Chen, Y., Rao, A., Jiang, X., Xiao, S., Ma, R., Wang, Z., Xiong, H., Dai, B.: Cinepregen: Camera controllable video previsualization via engine-powered diffusion. arXiv preprint arXiv:2408.17424  (2024)

\bibitem{christie2008camera}
Christie, M., Olivier, P., Normand, J.M.: Camera control in computer graphics. In: Computer graphics forum. vol.~27, pp. 2197--2218. Wiley Online Library (2008)

\bibitem{courant2024etexceptionaltrajectoriestexttocameratrajectory}
Courant, R., Dufour, N., Wang, X., Christie, M., Kalogeiton, V.: E.t. the exceptional trajectories: Text-to-camera-trajectory generation with character awareness (2024), \url{https://arxiv.org/abs/2407.01516}

\bibitem{courant2025pulp}
Courant, R., Wang, X., Loiseaux, D., Christie, M., Kalogeiton, V.: Pulp motion: Framing-aware multimodal camera and human motion generation. arXiv preprint arXiv:2510.05097  (2025)

\bibitem{evin2022cine}
Evin, I., H{\"a}m{\"a}l{\"a}inen, P., Guckelsberger, C.: Cine-ai: Generating video game cutscenes in the style of human directors. Proceedings of the ACM on Human-Computer Interaction  \textbf{6}(CHI PLAY),  1--23 (2022)

\bibitem{galvane2015automatic}
Galvane, Q.: Automatic cinematography and editing in virtual environments. Ph.D. thesis, Grenoble 1 UJF-Universit{\'e} Joseph Fourier (2015)

\bibitem{gschwindt2019can}
Gschwindt, M., Camci, E., Bonatti, R., Wang, W., Kayacan, E., Scherer, S.: Can a robot become a movie director? learning artistic principles for aerial cinematography. In: 2019 IEEE/RSJ International Conference on Intelligent Robots and Systems (IROS). pp. 1107--1114. IEEE (2019)

\bibitem{guo2023animatediff}
Guo, Y., Yang, C., Rao, A., Liang, Z., Wang, Y., Qiao, Y., Agrawala, M., Lin, D., Dai, B.: Animatediff: Animate your personalized text-to-image diffusion models without specific tuning. arXiv preprint arXiv:2307.04725  (2023)

\bibitem{hayashi2014t2v}
Hayashi, M., Inoue, S., Douke, M., Hamaguchi, N., Kaneko, H., Bachelder, S., Nakajima, M.: T2v: New technology of converting text to cg animation. ITE Transactions on Media Technology and Applications  \textbf{2}(1),  74--81 (2014)

\bibitem{he2024cameractrl}
He, H., Xu, Y., Guo, Y., Wetzstein, G., Dai, B., Li, H., Yang, C.: Cameractrl: Enabling camera control for text-to-video generation. arXiv preprint arXiv:2404.02101  (2024)

\bibitem{hu2024motionmaster}
Hu, T., Zhang, J., Yi, R., Wang, Y., Huang, H., Weng, J., Wang, Y., Ma, L.: Motionmaster: Training-free camera motion transfer for video generation (2024)

\bibitem{huang2021one}
Huang, C., Dang, Y., Chen, P., Yang, X., Cheng, K.T.: One-shot imitation drone filming of human motion videos. IEEE Transactions on Pattern Analysis and Machine Intelligence  \textbf{44}(9),  5335--5348 (2021)

\bibitem{huang2019learning}
Huang, C., Lin, C.E., Yang, Z., Kong, Y., Chen, P., Yang, X., Cheng, K.T.: Learning to film from professional human motion videos. In: Proceedings of the IEEE/CVF Conference on Computer Vision and Pattern Recognition. pp. 4244--4253 (2019)

\bibitem{huang2025patchdpo}
Huang, Q., Chan, L., Liu, J., He, W., Jiang, H., Song, M., Song, J.: Patchdpo: Patch-level dpo for finetuning-free personalized image generation. In: Proceedings of the Computer Vision and Pattern Recognition Conference. pp. 18369--18378 (2025)

\bibitem{huang2023vbenchcomprehensivebenchmarksuite}
Huang, Z., He, Y., Yu, J., Zhang, F., Si, C., Jiang, Y., Zhang, Y., Wu, T., Jin, Q., Chanpaisit, N., Wang, Y., Chen, X., Wang, L., Lin, D., Qiao, Y., Liu, Z.: Vbench: Comprehensive benchmark suite for video generative models (2023), \url{https://arxiv.org/abs/2311.17982}

\bibitem{jiang2021camera}
Jiang, H., Christie, M., Wang, X., Liu, L., Wang, B., Chen, B.: Camera keyframing with style and control. ACM Transactions on Graphics (TOG)  \textbf{40}(6),  1--13 (2021)

\bibitem{jiang2024cinematographic}
Jiang, H., Wang, X., Christie, M., Liu, L., Chen, B.: Cinematographic camera diffusion model. In: Computer Graphics Forum. vol.~43, p. e15055. Wiley Online Library (2024)

\bibitem{jiang2024cinematic}
Jiang, X., Rao, A., Wang, J., Lin, D., Dai, B.: Cinematic behavior transfer via nerf-based differentiable filming. In: Proceedings of the IEEE/CVF Conference on Computer Vision and Pattern Recognition. pp. 6723--6732 (2024)

\bibitem{vace}
Jiang, Z., Han, Z., Mao, C., Zhang, J., Pan, Y., Liu, Y.: Vace: All-in-one video creation and editing. In: Proceedings of the IEEE/CVF International Conference on Computer Vision. pp. 17191--17202 (2025)

\bibitem{karakostas2020shot}
Karakostas, I., Mademlis, I., Nikolaidis, N., Pitas, I.: Shot type constraints in uav cinematography for autonomous target tracking. Information Sciences  \textbf{506},  273--294 (2020)

\bibitem{kuang2024collaborative}
Kuang, Z., Cai, S., He, H., Xu, Y., Li, H., Guibas, L.J., Wetzstein, G.: Collaborative video diffusion: Consistent multi-video generation with camera control. Advances in Neural Information Processing Systems  \textbf{37},  16240--16271 (2024)

\bibitem{li2025veubenchcomprehensiveunderstandingvideo}
Li, B., Wu, Y., Lu, Y., Yu, J., Tang, L., Cao, J., Zhu, W., Sun, Y., Wu, J., Zhu, W.: Veu-bench: Towards comprehensive understanding of video editing (2025), \url{https://arxiv.org/abs/2504.17828}

\bibitem{li2024mvbenchcomprehensivemultimodalvideo}
Li, K., Wang, Y., He, Y., Li, Y., Wang, Y., Liu, Y., Wang, Z., Xu, J., Chen, G., Luo, P., Wang, L., Qiao, Y.: Mvbench: A comprehensive multi-modal video understanding benchmark (2024), \url{https://arxiv.org/abs/2311.17005}

\bibitem{li2025selfrewardingvisionlanguagemodelreasoning}
Li, Z., Yu, W., Huang, C., Liu, R., Liang, Z., Liu, F., Che, J., Yu, D., Boyd-Graber, J., Mi, H., Yu, D.: Self-rewarding vision-language model via reasoning decomposition (2025), \url{https://arxiv.org/abs/2508.19652}

\bibitem{liang2024richhumanfeedbacktexttoimage}
Liang, Y., He, J., Li, G., Li, P., Klimovskiy, A., Carolan, N., Sun, J., Pont-Tuset, J., Young, S., Yang, F., Ke, J., Dvijotham, K.D., Collins, K., Luo, Y., Li, Y., Kohlhoff, K.J., Ramachandran, D., Navalpakkam, V.: Rich human feedback for text-to-image generation (2024), \url{https://arxiv.org/abs/2312.10240}

\bibitem{lino2015intuitive}
Lino, C., Christie, M.: Intuitive and efficient camera control with the toric space. ACM Transactions on Graphics (TOG)  \textbf{34}(4),  1--12 (2015)

\bibitem{liu2025shotbenchexpertlevelcinematicunderstanding}
Liu, H., He, J., Jin, Y., Zheng, D., Dong, Y., Zhang, F., Huang, Z., He, Y., Li, Y., Chen, W., Qiao, Y., Ouyang, W., Zhao, S., Liu, Z.: Shotbench: Expert-level cinematic understanding in vision-language models (2025), \url{https://arxiv.org/abs/2506.21356}

\bibitem{liu2025improving}
Liu, J., Liu, G., Liang, J., Yuan, Z., Liu, X., Zheng, M., Wu, X., Wang, Q., Xia, M., Wang, X., et~al.: Improving video generation with human feedback. arXiv preprint arXiv:2501.13918  (2025)

\bibitem{liu2025videodpo}
Liu, R., Wu, H., Zheng, Z., Wei, C., He, Y., Pi, R., Chen, Q.: Videodpo: Omni-preference alignment for video diffusion generation. In: Proceedings of the Computer Vision and Pattern Recognition Conference. pp. 8009--8019 (2025)

\bibitem{liu2024videodpoomnipreferencealignmentvideo}
Liu, R., Wu, H., Ziqiang, Z., Wei, C., He, Y., Pi, R., Chen, Q.: Videodpo: Omni-preference alignment for video diffusion generation (2024), \url{https://arxiv.org/abs/2412.14167}

\bibitem{louarn2018automated}
Louarn, A., Christie, M., Lamarche, F.: Automated staging for virtual cinematography. In: Proceedings of the 11th ACM SIGGRAPH Conference on Motion, Interaction and Games. pp. 1--10 (2018)

\bibitem{lukasik2025better}
Lukasik, M., Meng, Z., Narasimhan, H., Chang, Y.W., Menon, A.K., Yu, F., Kumar, S.: Better autoregressive regression with llms via regression-aware fine-tuning. In: The Thirteenth International Conference on Learning Representations (2025)

\bibitem{lukasik2024regression}
Lukasik, M., Narasimhan, H., Menon, A.K., Yu, F., Kumar, S.: Regression-aware inference with llms. arXiv preprint arXiv:2403.04182  (2024)

\bibitem{maaz2024videochatgptdetailedvideounderstanding}
Maaz, M., Rasheed, H., Khan, S., Khan, F.S.: Video-chatgpt: Towards detailed video understanding via large vision and language models (2024), \url{https://arxiv.org/abs/2306.05424}

\bibitem{prabhudesai2024videodiffusionalignmentreward}
Prabhudesai, M., Mendonca, R., Qin, Z., Fragkiadaki, K., Pathak, D.: Video diffusion alignment via reward gradients (2024), \url{https://arxiv.org/abs/2407.08737}

\bibitem{pueyo2024cinempc}
Pueyo, P., Dendarieta, J., Montijano, E., Murillo, A.C., Schwager, M.: Cinempc: A fully autonomous drone cinematography system incorporating zoom, focus, pose, and scene composition. IEEE Transactions on Robotics  \textbf{40},  1740--1757 (2024)

\bibitem{rao2023dynamic}
Rao, A., Jiang, X., Guo, Y., Xu, L., Yang, L., Jin, L., Lin, D., Dai, B.: Dynamic storyboard generation in an engine-based virtual environment for video production. In: ACM SIGGRAPH 2023 Posters, pp.~1--2 (2023)

\bibitem{ren2025gen3c3dinformedworldconsistentvideo}
Ren, X., Shen, T., Huang, J., Ling, H., Lu, Y., Nimier-David, M., Müller, T., Keller, A., Fidler, S., Gao, J.: Gen3c: 3d-informed world-consistent video generation with precise camera control (2025), \url{https://arxiv.org/abs/2503.03751}

\bibitem{subramonyam2018taketoons}
Subramonyam, H., Li, W., Adar, E., Dontcheva, M.: Taketoons: Script-driven performance animation. In: Proceedings of the 31st annual ACM symposium on user interface software and technology. pp. 663--674 (2018)

\bibitem{tong2025delving}
Tong, C., Guo, Z., Zhang, R., Shan, W., Wei, X., Xing, Z., Li, H., Heng, P.A.: Delving into rl for image generation with cot: A study on dpo vs. grpo. arXiv preprint arXiv:2505.17017  (2025)

\bibitem{unity}
Unity: Real-time development platform (2025), \url{https://unity.com}

\bibitem{unreal}
Unreal: We make the engine. you make it unreal. (2025), \url{https://www.unrealengine.com/}

\bibitem{wang2025frameinnoutunboundedcontrollable}
Wang, B., Chen, X., Gadelha, M., Cheng, Z.: Frame in-n-out: Unbounded controllable image-to-video generation (2025), \url{https://arxiv.org/abs/2505.21491}

\bibitem{wang2024multilinguale5textembeddings}
Wang, L., Yang, N., Huang, X., Yang, L., Majumder, R., Wei, F.: Multilingual e5 text embeddings: A technical report. arXiv preprint arXiv:2402.05672  (2024)

\bibitem{wang2024qwen2vlenhancingvisionlanguagemodels}
Wang, P., Bai, S., Tan, S., Wang, S., Fan, Z., Bai, J., Chen, K., Liu, X., Wang, J., Ge, W., Fan, Y., Dang, K., Du, M., Ren, X., Men, R., Liu, D., Zhou, C., Zhou, J., Lin, J.: Qwen2-vl: Enhancing vision-language model's perception of the world at any resolution (2024), \url{https://arxiv.org/abs/2409.12191}

\bibitem{wang2025cinemaster3dawarecontrollableframework}
Wang, Q., Luo, Y., Shi, X., Jia, X., Lu, H., Xue, T., Wang, X., Wan, P., Zhang, D., Gai, K.: Cinemaster: A 3d-aware and controllable framework for cinematic text-to-video generation (2025), \url{https://arxiv.org/abs/2502.08639}

\bibitem{wang2024akira}
Wang, X., Courant, R., Christie, M., Kalogeiton, V.: Akira: Augmentation kit on rays for optical video generation. arXiv preprint arXiv:2412.14158  (2024)

\bibitem{wang2023jaws}
Wang, X., Courant, R., Shi, J., Marchand, E., Christie, M.: Jaws: Just a wild shot for cinematic transfer in neural radiance fields. In: Proceedings of the IEEE/CVF Conference on Computer Vision and Pattern Recognition. pp. 16933--16942 (2023)

\bibitem{wang2024motionctrl}
Wang, Z., Yuan, Z., Wang, X., Li, Y., Chen, T., Xia, M., Luo, P., Shan, Y.: Motionctrl: A unified and flexible motion controller for video generation. In: ACM SIGGRAPH 2024 Conference Papers. pp. 1--11 (2024)

\bibitem{wei2025cinevisioninteractiveprevisualizationstoryboard}
Wei, Z., Wu, H., Zhang, L., Xu, X., Zheng, Y., Hui, P., Agrawala, M., Qu, H., Rao, A.: Cinevision: An interactive pre-visualization storyboard system for director-cinematographer collaboration (2025), \url{https://arxiv.org/abs/2507.20355}

\bibitem{xie2023camera}
Xie, C., Hemmi, I., Shishido, H., Kitahara, I.: Camera motion generation method based on performer's position for performance filming. In: 2023 IEEE 12th Global Conference on Consumer Electronics (GCCE). pp. 957--960. IEEE (2023)

\bibitem{xing2025motioncanvascinematicshotdesign}
Xing, J., Mai, L., Ham, C., Huang, J., Mahapatra, A., Fu, C.W., Wong, T.T., Liu, F.: Motioncanvas: Cinematic shot design with controllable image-to-video generation (2025), \url{https://arxiv.org/abs/2502.04299}

\bibitem{xu2024filmagent}
Xu, Z., Wang, J., Wang, L., Li, Z., Shi, S., Hu, B., Zhang, M.: Filmagent: Automating virtual film production through a multi-agent collaborative framework. In: SIGGRAPH Asia 2024 Technical Communications, pp.~1--4 (2024)

\bibitem{yang2024direct}
Yang, S., Hou, L., Huang, H., Ma, C., Wan, P., Zhang, D., Chen, X., Liao, J.: Direct-a-video: Customized video generation with user-directed camera movement and object motion. In: ACM SIGGRAPH 2024 Conference Papers. pp. 1--12 (2024)

\bibitem{yu2023novel}
Yu, Z., Wang, H., Katsaggelos, A.K., Ren, J.: A novel automatic content generation and optimization framework. IEEE Internet of Things Journal  \textbf{10}(14),  12338--12351 (2023)

\bibitem{yu2023automated}
Yu, Z., Wu, X., Wang, H., Katsaggelos, A.K., Ren, J.: Automated adaptive cinematography for user interaction in open world. IEEE Transactions on Multimedia  \textbf{26},  6178--6190 (2023)

\bibitem{yu2022enabling}
Yu, Z., Yu, C., Wang, H., Ren, J.: Enabling automatic cinematography with reinforcement learning. In: 2022 IEEE 5th International Conference on Multimedia Information Processing and Retrieval (MIPR). pp. 103--108. IEEE (2022)

\bibitem{yuan2025newtongenphysicsconsistentcontrollabletexttovideo}
Yuan, Y., Wang, X., Wickremasinghe, T., Nadir, Z., Ma, B., Chan, S.H.: Newtongen: Physics-consistent and controllable text-to-video generation via neural newtonian dynamics (2025), \url{https://arxiv.org/abs/2509.21309}

\bibitem{zhang2025gendopautoregressivecameratrajectory}
Zhang, M., Wu, T., Tan, J., Liu, Z., Wetzstein, G., Lin, D.: Gendop: Auto-regressive camera trajectory generation as a director of photography (2025), \url{https://arxiv.org/abs/2504.07083}

\bibitem{zhang2024directpreferenceoptimizationvideo}
Zhang, R., Gui, L., Sun, Z., Feng, Y., Xu, K., Zhang, Y., Fu, D., Li, C., Hauptmann, A., Bisk, Y., Yang, Y.: Direct preference optimization of video large multimodal models from language model reward (2024), \url{https://arxiv.org/abs/2404.01258}

\bibitem{zhu2025soft}
Zhu, Y., Wang, X., Lathuili{\`e}re, S., Kalogeiton, V.: Soft-di [m] o: Improving one-step discrete image generation with soft embeddings. arXiv preprint arXiv:2509.22925  (2025)

\end{thebibliography}


\clearpage
\begin{center}
{\Large \bfseries\boldmath
\pretolerance=10000
Appendix to VERTIGO: Visual Preference Optimization for Cinematic Camera Generation\par}
\vskip .8cm
\end{center}

\appendix 

\section{Additional Related Work}
\label{sec:suppl_related_work}

\subsection{Camera Trajectory Generation}

Traditional approaches in automatic cinematography employ predefined rules and optimization techniques to control camera behavior. Early works utilized analytical methods based on script annotations to automate basic movements like pan, tilt, and zoom, ensuring subject visibility within frames~\cite{hayashi2014t2v, subramonyam2018taketoons}. Optimization approaches manage complex camera intrinsics (e.g., focal length~\cite{karakostas2020shot}) and extrinsics~\cite{louarn2018automated, galvane2015automatic}, integrating aesthetic principles like composition~\cite{pueyo2024cinempc}, viewpoint continuity~\cite{bonatti2020autonomous}, and action coherence~\cite{yu2023novel}. Recent advancements leverage neural networks for enhanced flexibility: reinforcement learning agents optimize camera trajectories via human preference scores~\cite{gschwindt2019can} or reward functions balancing aesthetics and fidelity~\cite{yu2022enabling}. Transformers~\cite{xie2023camera} and GAN architectures~\cite{yu2023automated} further enhance tracking precision and interactive scene adaptability. While reference-based methods~\cite{huang2019learning, huang2021one} and NeRF-based techniques~\cite{wang2023jaws, jiang2024cinematic} transfer shooting patterns by aligning human kinematics or heatmap guidance, they primarily focus on long-range aerial shots, lacking fine-grained artistic control. Recent works address this by learning cinematic features from film masterpieces, such as extracting pose-based patterns for camera behavior synthesis~\cite{huang2019learning} or introducing keyframe constraints for stylistic consistency~\cite{jiang2021camera}.

Emerging learning-based methods leverage diffusion models to generate plausible camera trajectories, yet struggle with multi-shot coherence and stylistic diversity~\cite{jiang2024cinematographic, courant2024etexceptionaltrajectoriestexttocameratrajectory}. E.T.~\cite{courant2024etexceptionaltrajectoriestexttocameratrajectory} focuses on autonomous camera movement but neglects critical cinematographic elements like shot scale, angle control, and multi-segment planning. Cine-AI~\cite{evin2022cine} leverages director-specific datasets to generate game cutscenes with style uniformity, combining user-adjustable storyboards and runtime automation in Unity. Pulp Motion~\cite{courant2025pulp} extends this by incorporating framing-aware multimodal camera and human motion generation, emphasizing compositional elements for improved immersion. Collectively, these approaches reduce manual adjustments but fail to holistically integrate narrative context with cinematographic principles, often overlooking emotional and spatial dynamics in virtual environments.

\subsection{Advanced Controllable Video Generation}

Controllable camera motion in video generation is critical for film and media production. Recent advancements have explored methodologies to enable user-directed camera and object motion control~\cite{guo2023animatediff, hu2024motionmaster}. However, these often rely on 2D-based frameworks, lacking explicit 3D spatial modeling, which leads to perspective inconsistencies and geometric implausibility. Methods like Direct-a-Video attempt to decouple object and camera motion via sparse spatial constraints but suffer from overlapping artifacts due to 2D attention mechanisms~\cite{yang2024direct}.

To address 3D-aware control, recent efforts condition generation on simplified 3D camera trajectories, omitting intrinsic parameters such as focal length~\cite{wang2024motionctrl, he2024cameractrl, kuang2024collaborative}. Existing frameworks universally neglect cinematographic principles and lack integration with film-script-driven multi-shot planning or established conventions, highlighting a gap between research and professional workflows. More recent efforts, such as GEN3C~\cite{ren2025gen3c3dinformedworldconsistentvideo}, introduce 3D-informed world-consistent video generation with precise camera control, leveraging NeRF-like representations for enhanced spatial coherence. CineMaster~\cite{wang2025cinemaster3dawarecontrollableframework} proposes a 3D-aware framework for cinematic text-to-video generation, incorporating controllable elements like camera paths and scene dynamics. MotionCanvas~\cite{xing2025motioncanvascinematicshotdesign} focuses on cinematic shot design with controllable image-to-video generation, enabling fine-grained motion customization. Frame In-N-Out~\cite{wang2025frameinnoutunboundedcontrollable} advances unbounded controllable image-to-video generation, supporting extended sequences with dynamic framing. NewtonGen~\cite{yuan2025newtongenphysicsconsistentcontrollabletexttovideo} emphasizes physics-consistent and controllable text-to-video generation, ensuring realistic motion adherence. These innovations aim to bridge the divide by unifying narrative intent, trajectory planning, and aesthetic-aware video synthesis, though challenges in multi-shot coherence and real-time applicability persist.
\section{Dataset}

Most existing 3D virtual camera trajectory datasets are reconstructed from real videos~\cite{courant2024etexceptionaltrajectoriestexttocameratrajectory, zhang2025gendopautoregressivecameratrajectory}, making precise cinematographic annotation difficult and limiting representation to basic motion patterns. Distinct from these datasets, our procedural synthesis framework in Unity fundamentally decouples camera motion from look-at constraints. This separation enables the generation of flexible paths that maintain precise geometric relationships with targets, allowing us to systematically cover complex cinematic compositions that are often underrepresented in real-world video collections.

\noindent
\textbf{Trajectory parameterization and augmentation.}
To ensure precise control over cinematic effects, the raw data in Unity is represented as $C^{\text{raw}}=\{t,q,f\}\in \mathbb{R}^{3+4+1}$, where $t$, $q$, and $f$ denote translation, quaternion rotation, and focal length, respectively. This representation supports advanced maneuvers like the dolly-zoom, where camera movement must be synchronized with lens adjustments to maintain composition. For instance, given a camera pose with an initial field of view (FoV) $f_1$, we adjust the target distance using Equation \ref{fovAugmentation} to strictly preserve the subject's framing size during translation:
\begin{equation}
d_2=d_1\cdot \frac{\tan(f_1/2)}{\tan(f_2/2)} \label{fovAugmentation}
\end{equation}
These raw parameters are subsequently serialized into a RealEstate10k-compatible format $\text{vec}(f, K, [R|t])$, ensuring compatibility with standard geometry-aware pipelines while retaining the procedural precision of the synthetic environment.

\noindent
\textbf{Automated Semantic Annotation.}
Leveraging the procedural nature of our generation pipeline, we eliminate the noise and labor associated with manual labeling. Because every trajectory is synthesized following specific compositional rules, ground-truth motion tags (e.g., Pan Left, Zoom In) are inherently linked to the data. To translate these structured attributes into the natural language required for VLM training, we employ \textit{Qwen-Plus} for large-scale batch inference. We utilize carefully engineered prompts to convert these categorical tags into coherent, descriptive captions. This automated process guarantees high-quality textual alignment for all 120K samples, directly supporting the semantic scoring mechanism described in the method section.

\noindent
\textbf{Cinematographic Taxonomy.}
Our annotation framework adopts a hierarchical cinematographic taxonomy spanning five core dimensions that align with professional filmmaking conventions. This structured classification enables precise control over both camera behavior and visual composition:

\begin{itemize}[leftmargin=*,itemsep=2pt,topsep=2pt]
\item \textbf{Shot Motion}: \textit{booms up/down, rotates, trucks, pushed in / out, zoom in / out, dolly zoom in / out, pan, tilt, static.}
\item \textbf{Shot Scale}: \textit{extreme close, close, medium close-up, medium, long, extreme long.}
\item \textbf{Shot Direction}: \textit{front, back, left, right, left front, right front, left back, right back.}
\item \textbf{Shot Angle}: \textit{high-angle, eye-level, low angle.}
\item \textbf{Screen Property}: \textit{up left, up center, up right, middle left, middle center, middle right, bottom left, bottom center, bottom right.}
\end{itemize}



\begin{table*}[t]
  \centering
  \footnotesize
  \setlength{\tabcolsep}{4pt}
  \caption{\textbf{Dataset comparison.} We compare the dataset with (i) RealEstate10k, a pure camera trajectory dataset designed for tasks like novel view synthesis; and (ii) three trajectory datasets captioned with detailed description on lens feature. Notice that Movement Types here only count basic motion types, excluding combinations of basic motions (as they can be easily achieved in game engines).}
  \resizebox{\textwidth}{!}{%
    \begin{tabular}{@{}l|l|ccc|cc|c@{}}
    \hline
    \multirow{2}[2]{*}{\textbf{Dataset}} & \multirow{2}[2]{*}{\textbf{Method}} & \multicolumn{3}{c|}{\textbf{Statistics}} & \multicolumn{2}{c|}{\textbf{Annotations}} & \textbf{Intrinsics} \\
          &       & \textbf{\#Samples} & \textbf{\#Frames} & \textbf{\#Movement} & \textbf{Caption} & \textbf{Tags} & \textbf{(FOV)} \\
    \hline
    RealEstate10k & SLAM / BA & 79K & 11M   & -     & \xmark & \xmark & \xmark \\
    E.T.  & SfM / HMR & 115K  & 11M   & 7     & \cmark & \xmark & \xmark \\
    DataDoP & SLAM  & 29K   & 11M   & 7     & \cmark & \xmark & \xmark \\
    CCD   & Synthetic & 25K   & 4.5M  & 10    & \cmark & \xmark & \xmark \\
    LenScript (Ours) & Synthetic & \best{120K} & \best{21.6M} & \best{13} & \cmark & \cmark & \cmark \\
    \hline
    \end{tabular}%
  }
  \label{data_statistics}%
\end{table*}%
\section{Application, Framework and Discussion}

\begin{figure*}[htbp]
  \centering
  \includegraphics[width=\linewidth]{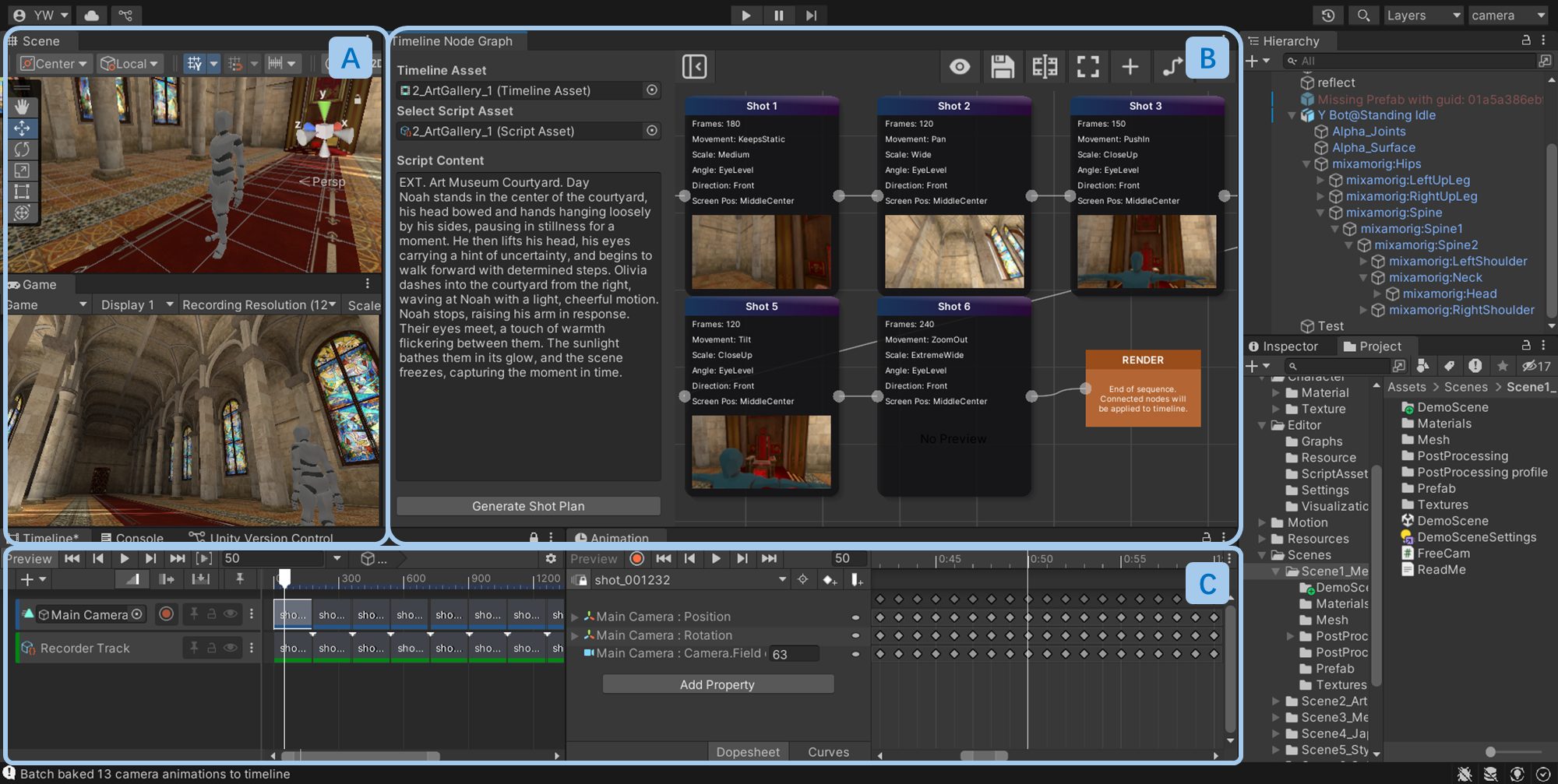}
  \caption{\textbf{Unity interface.} Panels: (A) 3D Scene and Game views for real-time asset layout and visual preview; (B) node-based storyboard with script editor and shot attribute panels for planning and prompt construction; (C) Timeline and Animation editors for fine-grained trajectory refinement, preview, and batch export supporting VLM scoring and DPO data collection.}
  \label{fig:unity_interface}
\end{figure*}

Our framework establishes a unified pipeline for virtual cinematography, integrating script parsing, camera planning, and preference optimization within Unity to enable efficient previsualization and iterative refinement, thereby bridging gaps between trajectory design and downstream video synthesis while accelerating artist workflows.

\textbf{Script Parsing and Camera Planning.} To derive cinematographically valid shot sequences from film scripts and virtual environments, we apply structured schema constraints $\mathbb{T}=\left\langle l_f, \Theta\right\rangle$, where $l_f\in \mathbb{R}^+$ denotes frame length, and $\Theta=\left\{t_i\right\}_{i=1}^6$ specifies cinematographic parameters under typological constraints:
\begin{equation}
t_i \in \bigcup_{j=1}^m V_j, \quad V_j=\left\{v_{j 1}, \ldots, v_{j m}\right\}
\end{equation}
with $V_j$ as film-theory-curated value sets following the cinematographic taxonomy. User-provided scripts are transformed into structured attributes, which are imported into the game engine and converted to natural-language prompts for guiding trajectory generation, producing pose sequences for video synthesis and preference learning.

\textbf{Scene-Aware Object Selection.} To align virtual environments with narrative contexts, we implement a layer-based selection mechanism that dynamically maps script entities to predefined game engine layers (e.g., characters, objects), ensuring coherent prioritization of shooting targets and scene compositions.

\textbf{Engine-based Workflow Integration.} Our system embeds trajectory generation and preference optimization directly within Unity, establishing an engine-native toolkit that supports seamless integration with industry-standard previsualization workflows. As illustrated in Fig.~\ref{fig:unity_interface}, the Unity interface comprises several interconnected components designed for intuitive artist interaction. (A) features the 3D scene view and game view, facilitating real-time organization of scene assets, such as props and characters, alongside immediate previews of shooting effects to ensure visual fidelity during planning. (B) introduces a node-based interactive interface, incorporating a film script editor for textual input and refinement, as well as shot attribute editors for specifying parameters like movement, scale, and angle. Users can organize shots into a storyboard via this nodal system, which automatically sequences them into a timeline. Subsequently, leveraging Unity's native timeline and animation editors, artists can perform fine-grained adjustments to virtual camera trajectories enabling precise control over motion dynamics without exiting the environment. We also provide a suite of tools for batch exporting rendered videos from Unity, trajectory previews, and other utilities to facilitate VLM scoring and DPO data construction. This integrated design not only aligns with established production practices used by technical directors and cinematographers but also provides real-time feedback loops, where generated trajectories can be previewed, edited, and exported alongside timeline utilities for Direct Preference Optimization (DPO) data collection, thereby minimizing workflow disruptions and enhancing iterative refinement.

\textbf{Downstream Applications.} By leveraging Unity's rapid rendering capabilities, our framework facilitates efficient previsualization that directly informs downstream video generation pipelines, enabling early validation of camera motions to reduce computational overhead in full synthesis. For instance, Unity-generated preview sequences can be employed in engine-powered diffusion models for cinematic previsualization and video generation~\cite{chen2024cinepregen}, allowing artists to refine trajectories before committing to resource-intensive rendering. Furthermore, the optimized camera trajectories serve as explicit conditioning signals during controllable text-to-video or image-to-video generation~\cite{he2024cameractrl, ren2025gen3c3dinformedworldconsistentvideo, wang2024akira}, enhancing geometric consistency and aesthetic quality in outputs from methods, thereby supporting scalable production workflows in virtual cinematography.
\section{User Study Details}

The main paper presents the best-of-4 results for Unity rendering and Wan 2.2 VACE video-to-video transfer. Here, we provide the questionnaire design and the full ranking results.

\subsection{Study Design}

\begin{figure}[t]
  \centering
  \includegraphics[width=\linewidth]{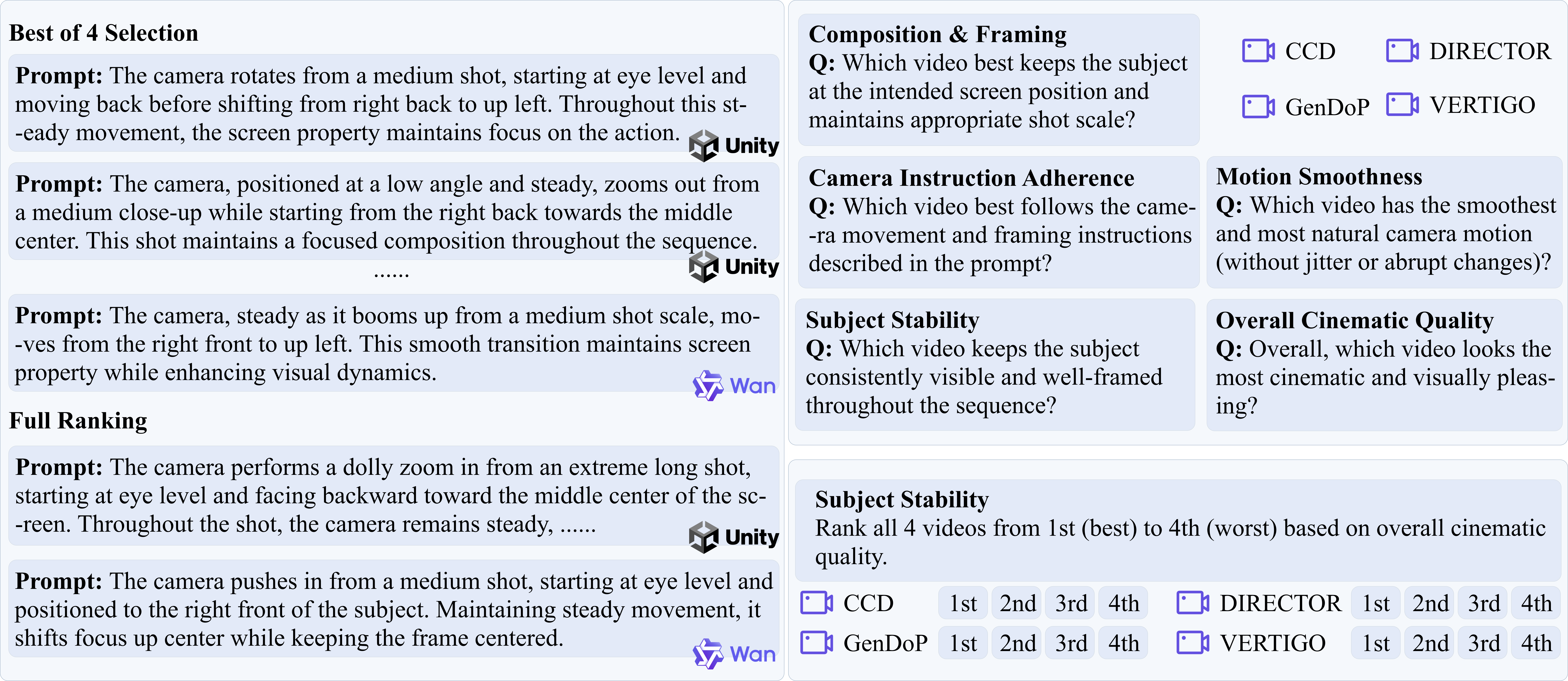}
  \caption{\textbf{Questionnaire interface.} Top: the best-of-4 interface shows the camera prompt together with five evaluation dimensions, where participants select the best result among four methods for each dimension. Bottom: the full-ranking interface asks participants to rank the four methods by overall cinematic quality.}
  \label{fig:questionnaire}
\end{figure}

To assess the perceived cinematic quality of generated trajectories, we collected 34 valid responses, including 11 expert participants with cinematography or video-production experience and 23 general viewers. As shown in Fig.~\ref{fig:questionnaire}, the questionnaire consists of two parts. In the \textit{best-of-4} task, participants view the camera prompt and choose the preferred result among CCD, DIRECTOR, GenDoP, and VERTIGO for five evaluation dimensions. In the \textit{full-ranking} task, participants rank the four methods by overall cinematic quality.

The study contains five best-of-4 groups and two full-ranking groups. Among them, three best-of-4 groups evaluate Unity-rendered trajectories and two evaluate Wan 2.2 VACE transfer results. For Unity, the five dimensions are composition \& framing, motion smoothness, camera instruction adherence, subject stability, and overall cinematic quality. For Wan 2.2 VACE, the five dimensions are composition \& framing, temporal consistency, content preservation, transfer quality, and freedom from artifacts.

\subsection{Best-of-4 Results}

VERTIGO achieves the highest vote share on every evaluated dimension across all five best-of-4 groups. Aggregating the three Unity groups, our method receives 46.1\% preference on composition, 50.0\% on motion smoothness, 53.9\% on instruction adherence, 49.0\% on subject stability, and 49.0\% on overall cinematic quality, yielding 49.6\% overall preference across all Unity questions. Aggregating the two Wan 2.2 VACE groups, VERTIGO obtains 60.3\% preference on composition, 55.9\% on temporal consistency, 57.4\% on content preservation, 55.9\% on transfer quality, and 52.9\% on artifact suppression, corresponding to 56.5\% overall preference. These results are consistent with the main paper and further verify that our preference post-training improves both trajectory-level control and downstream transfer robustness.

\subsection{Full-Ranking Results}

\begin{table*}[t]
  \centering
  \footnotesize
  \setlength{\tabcolsep}{5pt}
  \resizebox{\textwidth}{!}{%
  \begin{tabular}{l|ccccc|ccccc}
    \toprule
    & \multicolumn{5}{c|}{\textbf{Unity ranking}} & \multicolumn{5}{c}{\textbf{Video Gen ranking}} \\
    \textbf{Method} & Avg.$\downarrow$ & 1st & 2nd & 3rd & 4th & Avg.$\downarrow$ & 1st & 2nd & 3rd & 4th \\
    \hline
    CCD & 2.97 & 8.8 & 14.7 & 47.1 & 29.4 & 3.03 & 5.9 & 20.6 & 38.2 & 35.3 \\
    DIRECTOR & 3.27 & 6.1 & 12.1 & 30.3 & 51.5 & 3.21 & 8.8 & 11.8 & 29.4 & 50.0 \\
    GenDoP & 2.26 & 14.7 & 55.9 & 17.6 & 11.8 & 2.18 & 29.4 & 35.3 & 23.5 & 11.8 \\
    VERTIGO & \best{1.47} & \best{70.6} & 17.6 & 5.9 & 5.9 & \best{1.59} & \best{55.9} & 32.4 & 8.8 & 2.9 \\
    \bottomrule
  \end{tabular}%
  }
  \caption{\textbf{Full-ranking results.} Percentages indicate how often each method is assigned each rank. A lower average rank is better.}
  \label{tab:user_study_ranking}
\end{table*}

Table~\ref{tab:user_study_ranking} reveals a clear difference between the Unity and transfer ranking tasks. In the Unity ranking group, VERTIGO is the dominant choice, receiving 70.6\% first-place votes and the best average rank of 1.47, substantially outperforming GenDoP (2.26), CCD (2.97), and DIRECTOR (3.27).

In the Wan 2.2 VACE ranking group, GenDoP achieves the strongest overall ranking performance, receiving 55.9\% first-place votes and the best average rank of 1.59, while VERTIGO ranks second with 29.4\% first-place votes and an average rank of 2.18. Together with the best-of-4 results, this suggests that dimension-wise judgments and holistic ranking capture complementary aspects of transfer quality: VERTIGO is consistently preferred on targeted criteria such as composition, temporal stability, and artifact suppression, while the single holistic ranking example in VG7 is more favorable to GenDoP.
\section{Additional Qualitative Results}

\begin{figure*}[htbp]
  \centering
  \includegraphics[width=\linewidth,scale=1.00]{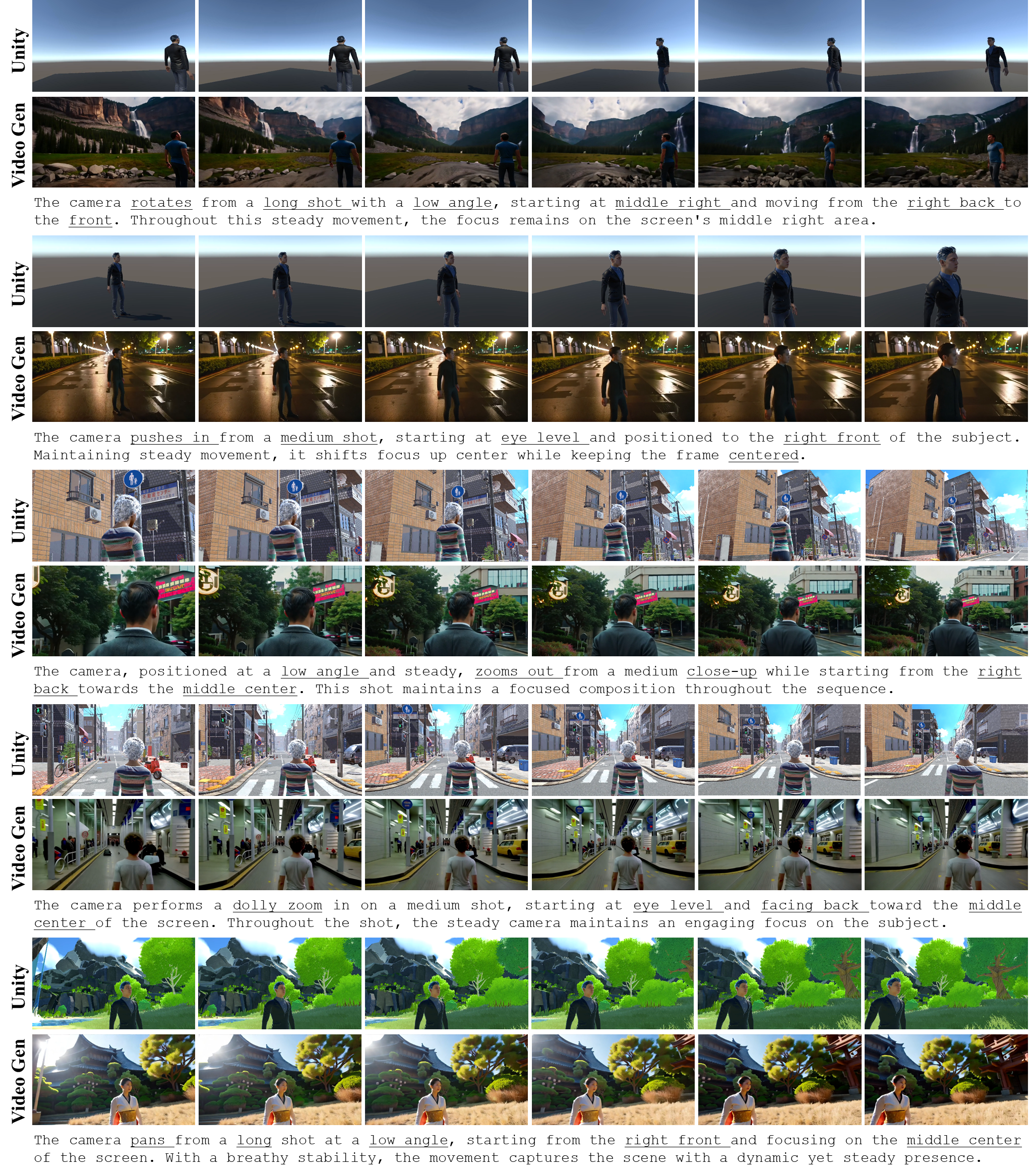}
  \caption{\textbf{Additional qualitative comparison of camera trajectories.} Across diverse scenes and prompts, VERTIGO maintains stable framing, consistent subject retention, and smooth motion while adhering to compositional intent.}	
  \label{fig:additional_qualitative}
\end{figure*}

\begin{figure*}[htbp]
  \centering
  \includegraphics[width=\linewidth]{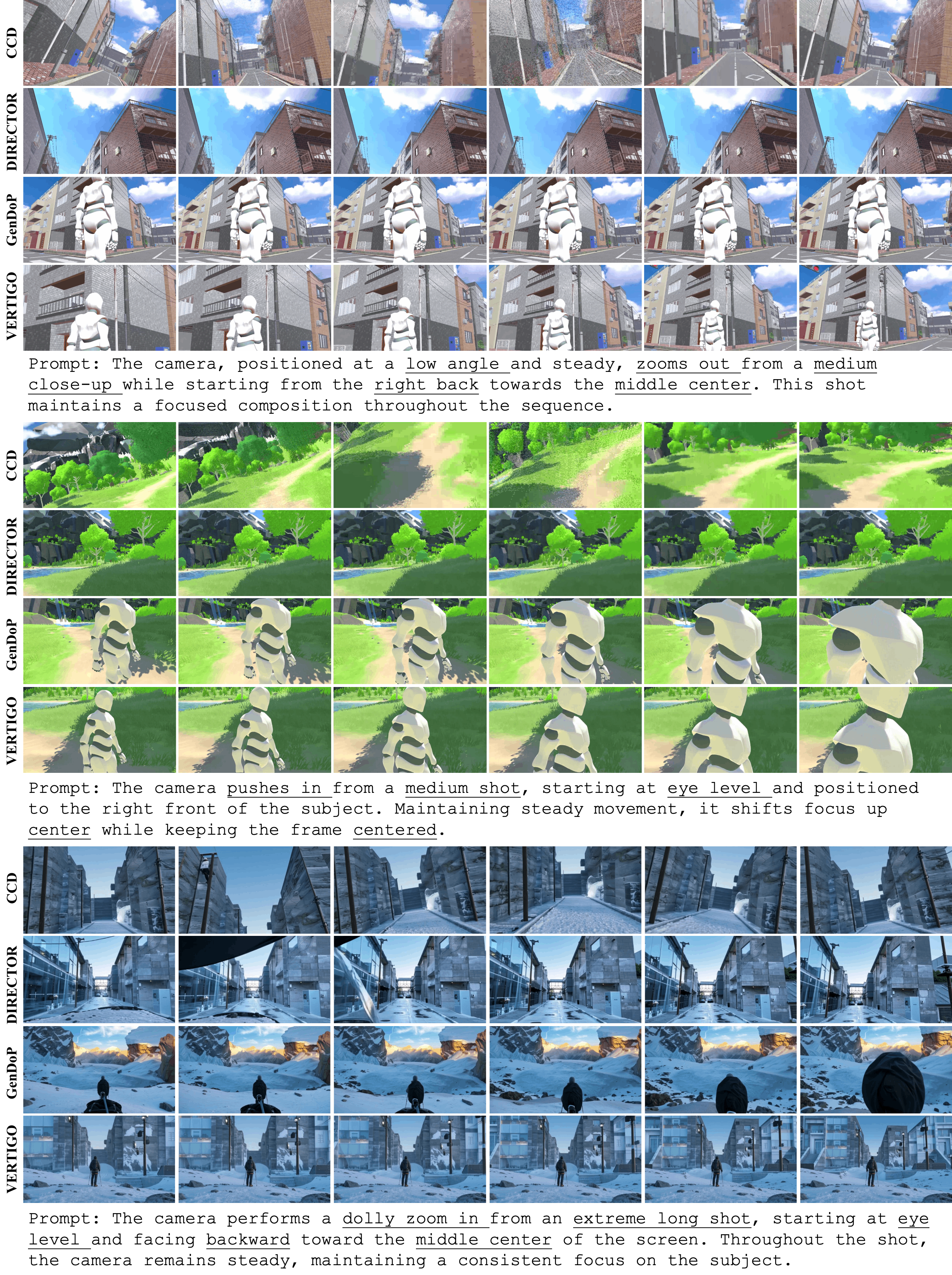}
  \caption{\textbf{Additional qualitative comparison in Unity rendering and video generation.} We compare CCD, DIRECTOR, GenDoP, and VERTIGO in both Unity-rendered previews and downstream video generation. VERTIGO produces more stable framing and stronger target retention in Unity, which transfers to more reliable composition and fewer visual failures in generated videos.}
  \label{fig:additional_qualitative_2}
\end{figure*}

We present additional qualitative examples across both trajectory-level and video-level comparisons. As illustrated in Fig.~\ref{fig:additional_qualitative}, VERTIGO generates smooth and cinematographically coherent camera motions that better preserve framing and compositional intent across diverse prompts and scenes. These results further demonstrate the robustness of our post-trained model for trajectory generation in Unity-based previsualization.

Figure~\ref{fig:additional_qualitative_2} further compares all four methods in both Unity rendering and downstream video generation. The improved framing behavior learned by VERTIGO is already visible in the Unity-rendered previews, where the subject remains better positioned and more consistently retained throughout the shot. This advantage transfers to video generation results, yielding more reliable composition, stronger temporal stability, and fewer framing failures than CCD, DIRECTOR, and GenDoP. Together, these examples support our key claim that lightweight render-in-the-loop preference optimization improves not only trajectory quality itself, but also the downstream visual quality of controllable video generation.

\end{document}